\PassOptionsToPackage{prologue,dvipsnames}{xcolor}
\documentclass{article}

\usepackage{microtype}
\usepackage{graphicx}
\usepackage{subcaption}
\usepackage{booktabs}
\usepackage{array}
\usepackage{multirow}
\usepackage{fvextra}
\usepackage{upquote}

\makeatletter
\def\PYG@reset{\let\PYG@it=\relax \let\PYG@bf=\relax%
    \let\PYG@ul=\relax \let\PYG@tc=\relax%
    \let\PYG@bc=\relax \let\PYG@ff=\relax}
\def\PYG@tok#1{\csname PYG@tok@#1\endcsname}
\def\PYG@toks#1+{\ifx\relax#1\empty\else%
    \PYG@tok{#1}\expandafter\PYG@toks\fi}
\def\PYG@do#1{\PYG@bc{\PYG@tc{\PYG@ul{%
    \PYG@it{\PYG@bf{\PYG@ff{#1}}}}}}}
\def\PYG#1#2{\PYG@reset\PYG@toks#1+\relax+\PYG@do{#2}}

\@namedef{PYG@tok@w}{\def\PYG@tc##1{\textcolor[rgb]{0.73,0.73,0.73}{##1}}}
\@namedef{PYG@tok@c}{\let\PYG@it=\textit\def\PYG@tc##1{\textcolor[rgb]{0.24,0.48,0.48}{##1}}}
\@namedef{PYG@tok@cp}{\def\PYG@tc##1{\textcolor[rgb]{0.61,0.40,0.00}{##1}}}
\@namedef{PYG@tok@k}{\let\PYG@bf=\textbf\def\PYG@tc##1{\textcolor[rgb]{0.00,0.50,0.00}{##1}}}
\@namedef{PYG@tok@kp}{\def\PYG@tc##1{\textcolor[rgb]{0.00,0.50,0.00}{##1}}}
\@namedef{PYG@tok@kt}{\def\PYG@tc##1{\textcolor[rgb]{0.69,0.00,0.25}{##1}}}
\@namedef{PYG@tok@o}{\def\PYG@tc##1{\textcolor[rgb]{0.40,0.40,0.40}{##1}}}
\@namedef{PYG@tok@ow}{\let\PYG@bf=\textbf\def\PYG@tc##1{\textcolor[rgb]{0.67,0.13,1.00}{##1}}}
\@namedef{PYG@tok@nb}{\def\PYG@tc##1{\textcolor[rgb]{0.00,0.50,0.00}{##1}}}
\@namedef{PYG@tok@nf}{\def\PYG@tc##1{\textcolor[rgb]{0.00,0.00,1.00}{##1}}}
\@namedef{PYG@tok@nc}{\let\PYG@bf=\textbf\def\PYG@tc##1{\textcolor[rgb]{0.00,0.00,1.00}{##1}}}
\@namedef{PYG@tok@nn}{\let\PYG@bf=\textbf\def\PYG@tc##1{\textcolor[rgb]{0.00,0.00,1.00}{##1}}}
\@namedef{PYG@tok@ne}{\let\PYG@bf=\textbf\def\PYG@tc##1{\textcolor[rgb]{0.80,0.25,0.22}{##1}}}
\@namedef{PYG@tok@nv}{\def\PYG@tc##1{\textcolor[rgb]{0.10,0.09,0.49}{##1}}}
\@namedef{PYG@tok@no}{\def\PYG@tc##1{\textcolor[rgb]{0.53,0.00,0.00}{##1}}}
\@namedef{PYG@tok@nl}{\def\PYG@tc##1{\textcolor[rgb]{0.46,0.46,0.00}{##1}}}
\@namedef{PYG@tok@ni}{\let\PYG@bf=\textbf\def\PYG@tc##1{\textcolor[rgb]{0.44,0.44,0.44}{##1}}}
\@namedef{PYG@tok@na}{\def\PYG@tc##1{\textcolor[rgb]{0.41,0.47,0.13}{##1}}}
\@namedef{PYG@tok@nt}{\let\PYG@bf=\textbf\def\PYG@tc##1{\textcolor[rgb]{0.00,0.50,0.00}{##1}}}
\@namedef{PYG@tok@nd}{\def\PYG@tc##1{\textcolor[rgb]{0.67,0.13,1.00}{##1}}}
\@namedef{PYG@tok@s}{\def\PYG@tc##1{\textcolor[rgb]{0.73,0.13,0.13}{##1}}}
\@namedef{PYG@tok@sd}{\let\PYG@it=\textit\def\PYG@tc##1{\textcolor[rgb]{0.73,0.13,0.13}{##1}}}
\@namedef{PYG@tok@si}{\let\PYG@bf=\textbf\def\PYG@tc##1{\textcolor[rgb]{0.64,0.35,0.47}{##1}}}
\@namedef{PYG@tok@se}{\let\PYG@bf=\textbf\def\PYG@tc##1{\textcolor[rgb]{0.67,0.36,0.12}{##1}}}
\@namedef{PYG@tok@sr}{\def\PYG@tc##1{\textcolor[rgb]{0.64,0.35,0.47}{##1}}}
\@namedef{PYG@tok@ss}{\def\PYG@tc##1{\textcolor[rgb]{0.10,0.09,0.49}{##1}}}
\@namedef{PYG@tok@sx}{\def\PYG@tc##1{\textcolor[rgb]{0.00,0.50,0.00}{##1}}}
\@namedef{PYG@tok@m}{\def\PYG@tc##1{\textcolor[rgb]{0.40,0.40,0.40}{##1}}}
\@namedef{PYG@tok@gh}{\let\PYG@bf=\textbf\def\PYG@tc##1{\textcolor[rgb]{0.00,0.00,0.50}{##1}}}
\@namedef{PYG@tok@gu}{\let\PYG@bf=\textbf\def\PYG@tc##1{\textcolor[rgb]{0.50,0.00,0.50}{##1}}}
\@namedef{PYG@tok@gd}{\def\PYG@tc##1{\textcolor[rgb]{0.63,0.00,0.00}{##1}}}
\@namedef{PYG@tok@gi}{\def\PYG@tc##1{\textcolor[rgb]{0.00,0.52,0.00}{##1}}}
\@namedef{PYG@tok@gr}{\def\PYG@tc##1{\textcolor[rgb]{0.89,0.00,0.00}{##1}}}
\@namedef{PYG@tok@ge}{\let\PYG@it=\textit}
\@namedef{PYG@tok@gs}{\let\PYG@bf=\textbf}
\@namedef{PYG@tok@gp}{\let\PYG@bf=\textbf\def\PYG@tc##1{\textcolor[rgb]{0.00,0.00,0.50}{##1}}}
\@namedef{PYG@tok@go}{\def\PYG@tc##1{\textcolor[rgb]{0.44,0.44,0.44}{##1}}}
\@namedef{PYG@tok@gt}{\def\PYG@tc##1{\textcolor[rgb]{0.00,0.27,0.87}{##1}}}
\@namedef{PYG@tok@err}{\def\PYG@bc##1{{\setlength{\fboxsep}{\string -\fboxrule}\fcolorbox[rgb]{1.00,0.00,0.00}{1,1,1}{\strut ##1}}}}
\@namedef{PYG@tok@kc}{\let\PYG@bf=\textbf\def\PYG@tc##1{\textcolor[rgb]{0.00,0.50,0.00}{##1}}}
\@namedef{PYG@tok@kd}{\let\PYG@bf=\textbf\def\PYG@tc##1{\textcolor[rgb]{0.00,0.50,0.00}{##1}}}
\@namedef{PYG@tok@kn}{\let\PYG@bf=\textbf\def\PYG@tc##1{\textcolor[rgb]{0.00,0.50,0.00}{##1}}}
\@namedef{PYG@tok@kr}{\let\PYG@bf=\textbf\def\PYG@tc##1{\textcolor[rgb]{0.00,0.50,0.00}{##1}}}
\@namedef{PYG@tok@bp}{\def\PYG@tc##1{\textcolor[rgb]{0.00,0.50,0.00}{##1}}}
\@namedef{PYG@tok@fm}{\def\PYG@tc##1{\textcolor[rgb]{0.00,0.00,1.00}{##1}}}
\@namedef{PYG@tok@vc}{\def\PYG@tc##1{\textcolor[rgb]{0.10,0.09,0.49}{##1}}}
\@namedef{PYG@tok@vg}{\def\PYG@tc##1{\textcolor[rgb]{0.10,0.09,0.49}{##1}}}
\@namedef{PYG@tok@vi}{\def\PYG@tc##1{\textcolor[rgb]{0.10,0.09,0.49}{##1}}}
\@namedef{PYG@tok@vm}{\def\PYG@tc##1{\textcolor[rgb]{0.10,0.09,0.49}{##1}}}
\@namedef{PYG@tok@sa}{\def\PYG@tc##1{\textcolor[rgb]{0.73,0.13,0.13}{##1}}}
\@namedef{PYG@tok@sb}{\def\PYG@tc##1{\textcolor[rgb]{0.73,0.13,0.13}{##1}}}
\@namedef{PYG@tok@sc}{\def\PYG@tc##1{\textcolor[rgb]{0.73,0.13,0.13}{##1}}}
\@namedef{PYG@tok@dl}{\def\PYG@tc##1{\textcolor[rgb]{0.73,0.13,0.13}{##1}}}
\@namedef{PYG@tok@s2}{\def\PYG@tc##1{\textcolor[rgb]{0.73,0.13,0.13}{##1}}}
\@namedef{PYG@tok@sh}{\def\PYG@tc##1{\textcolor[rgb]{0.73,0.13,0.13}{##1}}}
\@namedef{PYG@tok@s1}{\def\PYG@tc##1{\textcolor[rgb]{0.73,0.13,0.13}{##1}}}
\@namedef{PYG@tok@mb}{\def\PYG@tc##1{\textcolor[rgb]{0.40,0.40,0.40}{##1}}}
\@namedef{PYG@tok@mf}{\def\PYG@tc##1{\textcolor[rgb]{0.40,0.40,0.40}{##1}}}
\@namedef{PYG@tok@mh}{\def\PYG@tc##1{\textcolor[rgb]{0.40,0.40,0.40}{##1}}}
\@namedef{PYG@tok@mi}{\def\PYG@tc##1{\textcolor[rgb]{0.40,0.40,0.40}{##1}}}
\@namedef{PYG@tok@il}{\def\PYG@tc##1{\textcolor[rgb]{0.40,0.40,0.40}{##1}}}
\@namedef{PYG@tok@mo}{\def\PYG@tc##1{\textcolor[rgb]{0.40,0.40,0.40}{##1}}}
\@namedef{PYG@tok@ch}{\let\PYG@it=\textit\def\PYG@tc##1{\textcolor[rgb]{0.24,0.48,0.48}{##1}}}
\@namedef{PYG@tok@cm}{\let\PYG@it=\textit\def\PYG@tc##1{\textcolor[rgb]{0.24,0.48,0.48}{##1}}}
\@namedef{PYG@tok@cpf}{\let\PYG@it=\textit\def\PYG@tc##1{\textcolor[rgb]{0.24,0.48,0.48}{##1}}}
\@namedef{PYG@tok@c1}{\let\PYG@it=\textit\def\PYG@tc##1{\textcolor[rgb]{0.24,0.48,0.48}{##1}}}
\@namedef{PYG@tok@cs}{\let\PYG@it=\textit\def\PYG@tc##1{\textcolor[rgb]{0.24,0.48,0.48}{##1}}}

\def\PYGZus{\char`\_}

\def\PYGZlt{\char`\<}
\def\PYGZgt{\char`\>}
\def\PYGZsh{\char`\#}

\def\PYGZhy{\char`\-}
\def\PYGZsq{\textquotesingle}

\makeatother

\usepackage{pifont}
\usepackage{stfloats}

\usepackage{hyperref}

\usepackage[accepted]{icml2027}
\newif\ifarxiv
\arxivtrue
\ifarxiv
\hypersetup{pdfsubject={}}
\fi
\usepackage{amsmath}
\usepackage{amssymb}
\usepackage{mathtools}
\usepackage{amsthm}

\usepackage[capitalize,noabbrev]{cleveref}

\theoremstyle{plain}
\newtheorem{theorem}{Theorem}[section]
\newtheorem{proposition}[theorem]{Proposition}
\newtheorem{lemma}[theorem]{Lemma}

\theoremstyle{definition}
\newtheorem{definition}[theorem]{Definition}

\theoremstyle{remark}

\usepackage[textsize=tiny]{todonotes}

\usepackage[OT1]{fontenc}

\usepackage[dvipsnames]{xcolor}
\usepackage[scr=boondoxo,
            ]{mathalpha}
\usepackage[mathcal]{euscript}
\usepackage{makecell}

\DeclareMathAlphabet{\mathbbb}{U}{bbold}{m}{n}
\newcommand{\x}{\times}
\newcommand{\X}{\ifmmode\times\else\texttimes\fi}
\newcommand{\R}{\mathbb{R}}

\newcommand{\T}{\top}

\newcommand{\op}[1] {\operatorname{#1}}
\newcommand{\B}[1] {\textbf{#1}}

\usepackage{textcomp}
\newcommand{\tildemid}{\raisebox{0.5ex}{\texttildelow}}

\usepackage{soul}

\definecolor{hdcol}{HTML}{E00000}
\definecolor{lycol}{HTML}{0000E0}
\definecolor{tycol}{HTML}{7F00FF}
\newcommand{\hdi}{\textcolor{hdcol}{\boldsymbol{i}}}
\newcommand{\hdj}{\textcolor{hdcol}{\boldsymbol{j}}}
\newcommand{\hli}{\textcolor{lycol}{l_i}}
\newcommand{\hlj}{\textcolor{lycol}{l_j}}
\newcommand{\hs}{\textcolor{tycol}{s}}

\icmltitlerunning{Rewirable Circuits in Transformers with Loom Attention}

\begin{document}

\twocolumn[
\icmltitle{Rewirable Circuits in Transformers with Loom Attention}

\icmlsetsymbol{equal}{*}

\begin{icmlauthorlist}
\icmlauthor{Da Xiao}{yyy}
\icmlauthor{Qingye Meng}{comp}
\icmlauthor{Shengping Li}{comp}
\icmlauthor{Xingyuan Yuan}{comp}
\end{icmlauthorlist}

\icmlaffiliation{yyy}{Beijing University of Posts and Telecommunications, Beijing, China}
\icmlaffiliation{comp}{ColorfulClouds Technology Co., Ltd., Beijing, China}

\icmlcorrespondingauthor{Da Xiao}{xiaoda99@bupt.edu.cn}

\icmlkeywords{Multi-Head Attention, Transformer}

\ifarxiv
\begin{center}
\vskip 0.04in
{\normalsize \textsuperscript{1}Beijing University of Posts and Telecommunications \quad \textsuperscript{2}ColorfulClouds Technology}\par\vspace{3pt}
{\small \texttt{xiaoda99@bupt.edu.cn} \quad \textbf{Code:}~\href{https://github.com/Caiyun-AI/LoomFormer}{\texttt{https://github.com/Caiyun-AI/LoomFormer}}}
\end{center}
\vskip 0.2in
\else
\vskip 0.3in
\fi
]

\ifarxiv
\makeatletter\global\icml@noticeprintedtrue\makeatother
\else
\printAffiliationsAndNotice{}
\fi

\begin{abstract}
We propose loom attention, which allows a Transformer to dynamically rewire its circuits at every token by interweaving a head's vertical attention over the outputs of earlier heads (to obtain its query, key and value) with its horizontal attention over tokens.
We instantiate vertical attention efficiently with normalized and gated linear attention over a per-token matrix memory, which stores the compressed head outputs and serves as a second residual stream.
This yields LoomFormer, a dual-stream architecture that splits control and content between the vector and matrix streams.
Across model sizes and training token budgets in language modeling and downstream tasks, LoomFormer matches Transformers trained with over \mbox{3\X} compute and recent residual-stream baselines (HC/mHC, MUDDFormer, AttnRes) trained with \mbox{\tildemid1.7--2.8\X}.
For example, trained on the same 300B tokens with only 2\% extra FLOPs, LoomPythia-2.8B achieves \tildemid1.6\X\ the average downstream gain of mHC-Pythia-2.8B over Pythia-2.8B, and matches (0-shot) or surpasses (5-shot) Pythia-12B.

\end{abstract}

\section{Introduction} \label{intro}
\begin{figure}[!t]
\centering
\includegraphics[width=\columnwidth]{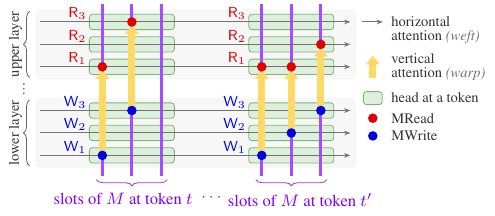}
\vskip -0.06in
\caption{Conceptual sketch of loom attention. At each token, each head chooses which slots of that token's matrix memory $M$ to read and write (i.e., vertical attention, every head does both; only one is shown per layer) and which tokens to attend to (horizontal attention), so a different circuit is woven for each token.}
\label{fig:weave}
\vskip 0.09in
\includegraphics[width=0.9\columnwidth]{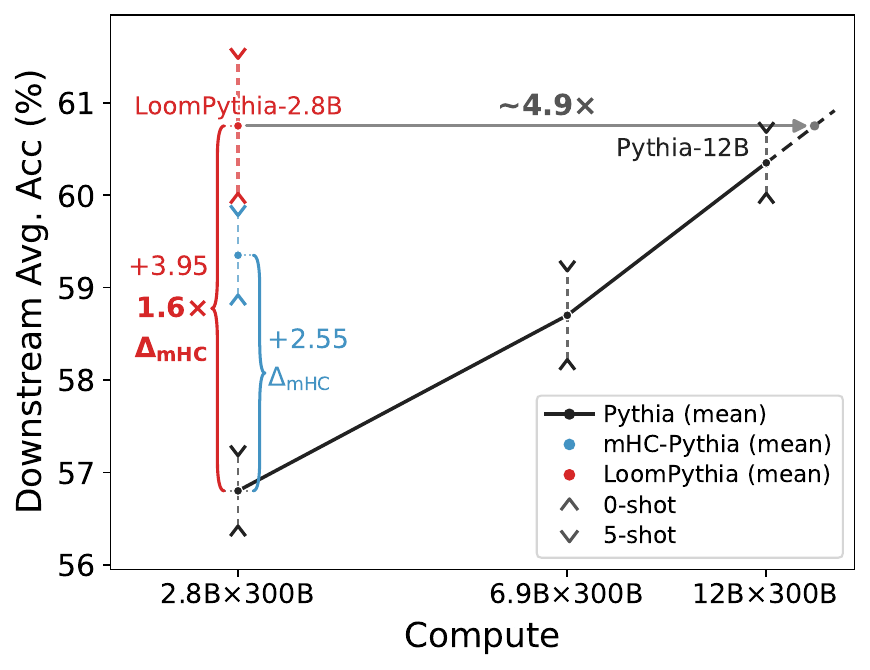}
\vskip -0.09in
\caption{Downstream accuracy averaged over 10 tasks.}%
\label{fig:loompythia-downstream}
\vskip -0.12in
\end{figure}
Transformers solve tasks by composing attention heads across layers into circuits \cite{Elhage2021mathematical,olsson2022context,wang2211interpretability,merullo2024talking}:
an earlier head writes a feature into a subspace of the residual stream, and a later head reads from that subspace into its query, key or value.
Induction heads, for example, read the output of a previous-token head through their keys (K-composition) \cite{olsson2022context}.

The composition between two heads, however, is determined by the product of their output and input projections, a \emph{virtual weight} that does not change after training \cite{Elhage2021mathematical}.
Attention makes the \emph{horizontal} flow of information across tokens input-dependent: a head chooses \emph{which tokens} to attend to.
In contrast, the \emph{vertical} flow across layers is fixed: a head cannot choose \emph{which heads} to read from.
A model that needs different circuits for different inputs therefore keeps all of them wired at once, lets them compete for the residual stream, and can gate them only indirectly, e.g., by sending a head's attention to a sink token \cite{xiao2023efficient}.
We ask whether the wiring itself can be chosen per token, i.e., whether circuits can be made \emph{rewirable}.

Recent work has enriched the vertical flow.
Dense connections let a layer aggregate the outputs of all earlier layers \cite{pagliardini2024denseformer};
MUDDFormer \cite{xiao2025muddformer} and Attention Residuals (AttnRes) \cite{kimi2026attention} make this aggregation input-dependent, turning it into a depth-wise (i.e., vertical) attention (separate for the query, key, value and residual inputs in MUDDFormer);
and Hyper-Connections (HC) \cite{zhu2024hyper} and its manifold-constrained variant mHC \cite{xie2025mhc} maintain several residual streams with learnable or dynamic mixing.
In all of them, however, the vertical channel meets horizontal attention only through the hidden state of a layer, in which the outputs of all heads have already been summed and mixed together:
a head can neither single out a particular earlier head nor be singled out by a later one.
The vertical and horizontal flows run side by side, never woven together.

To make circuits rewirable, we propose \emph{loom attention}, in which a head chooses which earlier heads to read from just as it chooses which tokens to attend to.
By attending vertically over the outputs of these heads at the same token, the head obtains its query, key and value to perform the standard horizontal attention over tokens, and its output is in turn attended to vertically by later heads.
Vertical and horizontal attention are thus interwoven, each feeding the other at every head, like the warp and weft of a loom (\cref{fig:weave}).

Naive vertical attention over the outputs of all earlier heads is, however, expensive.
We therefore instantiate it with normalized and gated linear attention over a per-token matrix memory, which stores the compressed head outputs and serves as a second residual stream.
We call the resulting dual-stream architecture \emph{LoomFormer}, after a drawloom---a loom in which a separate stored pattern selects which warp threads to lift at each pick, rather than repeating one fixed pattern.
\footnote{The automated form of the drawloom, the Jacquard loom, encoded this pattern on punched cards and inspired early programmable computers such as Babbage's Analytical Engine.}
Here the ordinary vector stream plays the pattern, carrying control---which memory slots each head writes and reads (weaving a different circuit per token) as well as which tokens to attend to---while the matrix stream carries the content moved across tokens.
The two streams are coupled through attention outputs and FFN writes.

Extensive experiments on language modeling and downstream tasks across model sizes and training token budgets show that LoomFormer adds only 2.0--3.7\% FLOPs while matching Transformers trained with over \mbox{3\X} compute and recent residual-stream baselines (HC/mHC, MUDDFormer, AttnRes) trained with \mbox{\tildemid1.7--2.8\X}.
For example, trained on the same 300B tokens, LoomPythia-2.8B achieves \tildemid1.6\X\ the average downstream gain of mHC-Pythia-2.8B over Pythia-2.8B, and matches (0-shot) or surpasses (5-shot) Pythia-12B (\cref{fig:loompythia-downstream}).
We also analyze the trained 2.8B models to understand how and why LoomFormer works---e.g., by rewiring head outputs to inputs per token, it keeps over three times as many deep-layer heads off attention sinks as Pythia and mHC.

\section{Rewirable Circuits} \label{framework}

\paragraph{Writers, readers and circuits.}
Attention heads communicate through the residual stream \cite{Elhage2021mathematical}.
Consider a Transformer with $L$ layers of $n$ heads of dim $d$, and model dim $D = nd$.
We call the output of a head a \emph{writer} and each of its inputs (query, key and value) a \emph{reader}, so a head has one writer and three readers.
Writer $\mathsf{W}_i$ adds its output $o_i \in \R^d$ to the residual stream through its block $W^O_i \in \R^{d \x D}$ of the output projection, and reader $\mathsf{R}_j$ of a later head reads the stream through its input projection $W^s_j \in \R^{D \x d}$ ($s \in \{Q, K, V\}$), so $\mathsf{W}_i$ reaches $\mathsf{R}_j$ through the \emph{virtual weight} $\bar C_{ij} = W^O_i W^s_j \in \R^{d \x d}$: the two communicate only if $\mathsf{W}_i$ writes into a subspace that $\mathsf{R}_j$ reads.
Heads composed in this way form \emph{circuits} whose nodes are the writers and readers.

\begin{figure}[t]
\centering
\includegraphics[width=0.9\columnwidth]{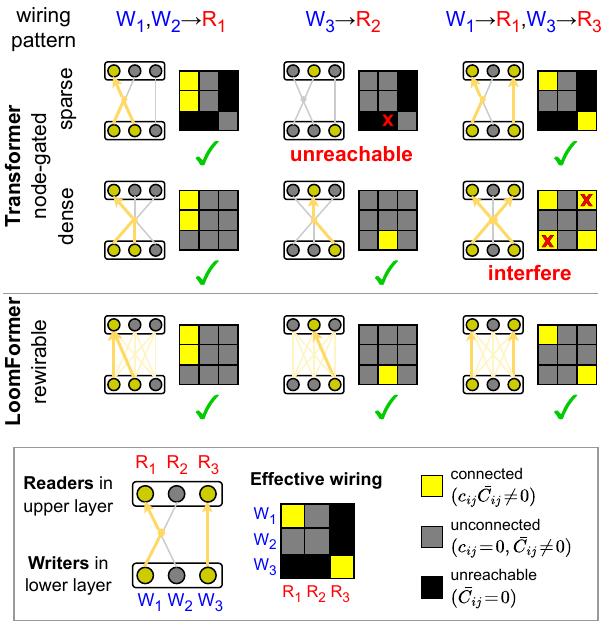}
\vskip -0.05in
\caption{Effective wiring for three wiring patterns.
Transformers (top two rows) can only gate whole heads, i.e., rows and columns, on and off, so they connect only rectangles within the fixed support of their virtual weights.
LoomFormer (bottom row) connects exactly the required pairs.}
\label{fig:rewire}
\vskip -0.1in
\end{figure}

A task may need different circuits at different tokens, i.e., different pairs of writers and readers to be connected (\cref{fig:rewire}).
To describe how the wiring can change, we give it a dynamic part besides the static virtual weights: at a token, reader $\mathsf{R}_j$ receives
\begin{equation} \label{eq:wiring}
    u_j = \textstyle\sum_i o_i\, c_{ij} \bar C_{ij},
\end{equation}
where the per-token scalars $c_{ij}$ form the \emph{wiring matrix} $C \in \R^{w \x r}$ from the $w$ earlier writers to the $r$ readers of one layer, and $c_{ij} \bar C_{ij}$ is the \emph{effective wiring} from $\mathsf{W}_i$ to $\mathsf{R}_j$.
The two are connected at the token if their effective wiring is nonzero.\footnote{In trained models entries are rarely exactly zero; we count those below a threshold as zero (\cref{app:bounds}).}
A \emph{wiring pattern} $P \subseteq [w] \x [r]$ is a set of writer-reader pairs, realized when each reader in $P$ is connected to exactly its writers in $P$.
\begin{definition} \label{def:rewirable}
An architecture with fixed virtual weights is \emph{rewirable w.r.t.} a set of wiring patterns if every pattern in the set is realized by some wiring matrix $C$ that it admits at a token.
\end{definition}

\paragraph{Transformers are node-gated.}
In a Transformer, pre-LN only rescales all writers of a token by a common factor (\cref{lemma:fixed} in \cref{app:node-gated}).
Within a token, nodes can thus only be gated through whole heads: a head that attends to a sink token \cite{xiao2023efficient} outputs nearly nothing, turning off both its writer and, since what its readers receive has no effect, its readers.
\begin{definition} \label{def:node-gated}
An architecture is \emph{node-gated} if, at every token, its wiring matrix has the form
\begin{equation} \label{eq:node-gated}
    C = g\, g'^\T,
\end{equation}
with gate vectors $g \in \R^{w}$ and $g' \in \R^{r}$ whose nonzero entries mark the writers and readers that are on.
\end{definition}
We call the rank of $C$ the \emph{wiring rank}; a node-gated architecture has wiring rank at most one, so existing connections can be gated on and off, but not rewired.
\begin{proposition} \label{prop:rectangle}
A Transformer with pre-LN is node-gated: at any token, it connects precisely the pairs with $\bar C_{ij} \ne 0$ whose writer and reader are both on, a rectangle within the fixed support of the virtual weights.
\end{proposition}
Gating a writer or a reader acts on its whole row or column of $C$, so a wiring pattern is realized only if it is such a rectangle, as in the first column of \cref{fig:rewire}; other patterns fail in one of two ways.
With \emph{dense} virtual weights, where heads share subspaces, every pair can be connected, but a non-rectangular pattern such as $\{\mathsf{W}_1 \to \mathsf{R}_1, \mathsf{W}_3 \to \mathsf{R}_3\}$ (last column) also connects $\mathsf{W}_1 \to \mathsf{R}_3$ and $\mathsf{W}_3 \to \mathsf{R}_1$: the two circuits \emph{interfere}.
With \emph{sparse} virtual weights, where groups of heads use separate subspaces, circuits in different groups do not interfere, but pairs across groups such as $\mathsf{W}_3 \to \mathsf{R}_2$ (second column) are \emph{unreachable}.

\paragraph{The price of rewiring by node gating.}
Can a node-gated architecture overcome this by spending more heads?
Suppose the role of each of $r \ge 2$ readers may be played by several heads, gated on and off per token, and at every token each reader must receive exactly one of $w$ writers, chosen independently of the other readers; the last pattern in \cref{fig:rewire} is the smallest instance.
Let $N(w, t)$ denote the minimum size of a ground set carrying a $t$-cover-free family of $w$ sets, i.e., $w$ sets none of which is covered by the union of $t$ others \cite{kautz1964nonrandom}.
\begin{theorem} \label{thm:cff}
To realize all such wiring patterns, a node-gated architecture needs at least $N(w, r-1)$ heads per reader, and this number suffices, where $N(w, 1) \approx \log_2 w$, $N(w, t) = \Omega(t^2 \log w / \log t)$ for $2 \le t \le \sqrt{w}$ and $N(w, t) = \Theta(w)$ for $t > \sqrt{w}$.
\end{theorem}
Intuitively, a head receives every writer that is on and has a nonzero virtual weight to it, so a reader must hold in reserve a head that sees its chosen writer and none of the up to $r-1$ writers chosen by the other readers (\cref{thm:cff-full} in \cref{app:cff}; bounds in \cref{app:bounds}).
With the $r = 3n$ readers of a layer choosing among its $w = nl$ earlier writers, $r - 1 > \sqrt{w}$ whenever $l \le 9n - 6$, so at practical sizes node gating needs $\Theta(w)$ heads per reader.

\paragraph{Naive loom attention makes Transformers rewirable.}
To rewire with one head per reader, the wiring matrix must be chosen entry by entry at each token.
The most direct way is to let every reader attend over the outputs of all earlier heads of the same token, with attention weights generated from the token itself:
\begin{equation} \label{eq:vertical}
    u_j = \textstyle\sum_i \alpha_{ij}(x)\, o_i, \qquad \alpha_{\cdot j}(x) = \op{softmax}(x W^\alpha_j),
\end{equation}
where $W^\alpha_j \in \R^{D \x nl}$ for a reader in layer $l$, which has $nl$ earlier writers.
We call this \emph{naive loom attention}: its virtual weights are identities and all the wiring is left to $C = [\alpha_{ij}]$, unlike a Transformer, which wires heads through fixed virtual weights and only gates them through a rank-one $C$.
With one head per reader, it is rewirable w.r.t.\ all wiring patterns (cf.\ \cref{fig:rewire}, bottom row), but expensive: over the $3n$ readers of all $L$ layers, generating the attention weights takes about $\frac{3}{2} n^2 L^2 D$ parameters and multiply-adds per token, growing quadratically in the total number of heads $nL$ and rivaling the whole model at practical sizes (\cref{app:cost}).

\section{Method} \label{method}
\begin{figure*}[t]
\centering
\includegraphics[width=0.95\textwidth]{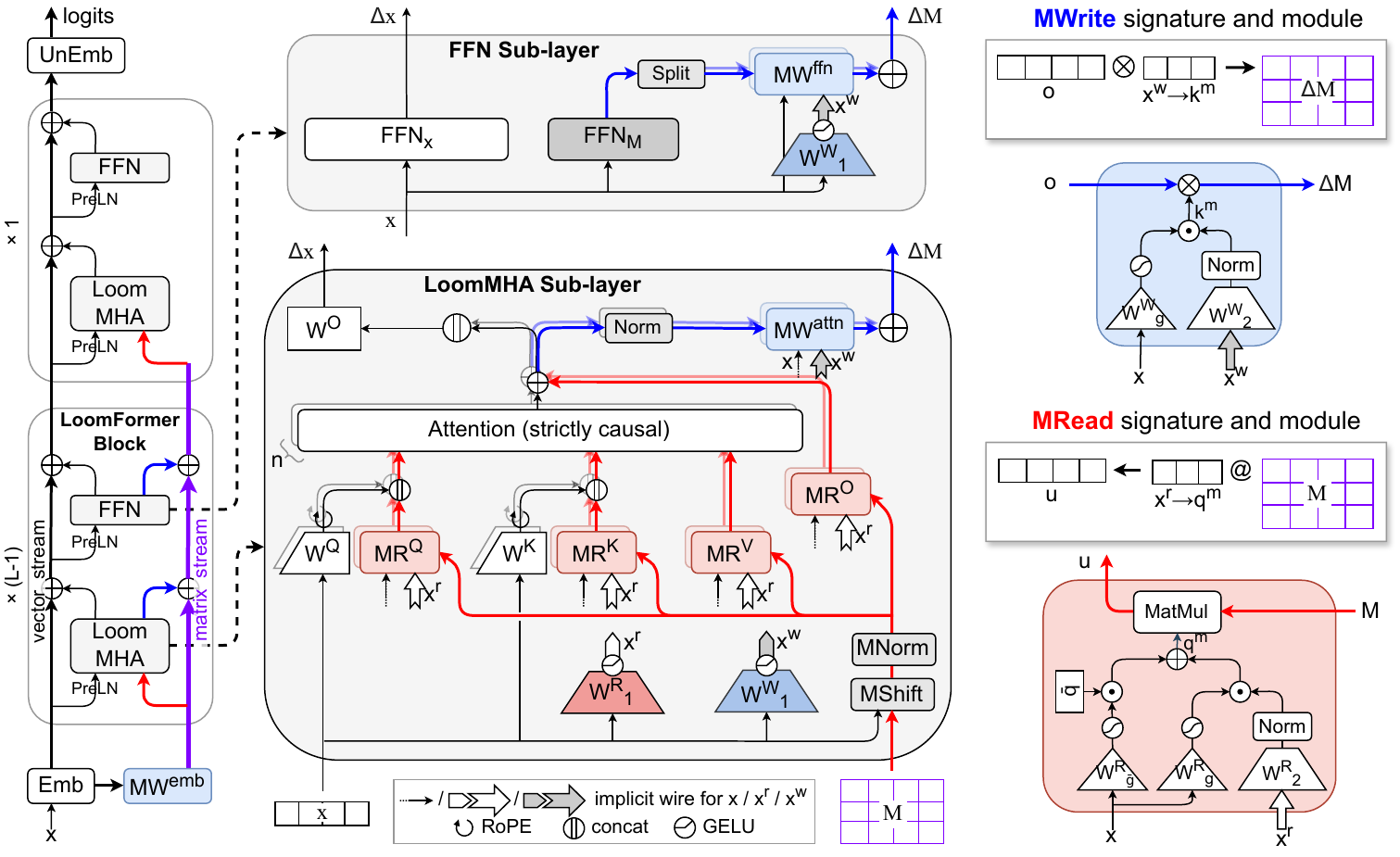}
\vskip -0.05in
\caption{LoomFormer architecture: the model (left), the LoomMHA and FFN sub-layers of a block (middle), and the MRead (MR) and MWrite (MW) modules (right).
Vertical attention: reads from $M$ are in red and writes to $M$ in blue.}
\label{fig:loom-arch}
\vskip -0.1in
\end{figure*}
\subsection{From Naive to Affordable Loom Attention} \label{affordable}
As linear attention compresses a growing key-value cache into a fixed-size state \cite{katharopoulos2020transformers,schlag2021linear}, we compress the head outputs of each token along depth into a matrix memory $M \in \R^{d_k \x d}$ of $d_k$ slots of width $d$.
Writer $\mathsf{W}_i$ writes its output to $M$ with a key $k^m_{i} \in \R^{d_k}$, and reader $\mathsf{R}_j$ reads with a query $q^m_{j} \in \R^{d_k}$, so a reader attends over $d_k$ slots instead of all earlier writers, at a cost independent of depth.
Since $M$ accumulates writes layer by layer, it is a second residual stream, the \emph{matrix stream}, besides the original \emph{vector stream} (\cref{fig:loom-arch} left).

We call the interface through which a head reads from or writes to a stream a \emph{port}.
The two streams have dual ports:
\begin{equation} \label{eq:port-duality}
\begin{split}
    & \textrm{vector stream: } u_j = x W^s_j, \quad \Delta x = o_i W^O_i \\
    & \textrm{matrix stream: } u_j = q^m_{j} M, \quad \Delta M = k^m_{i} \otimes o_i
\end{split}
\end{equation}
where $k^m_{i} \otimes o_i = (k^m_{i})^\T o_i$ is the outer product.
Data and ports swap shapes: a port shrinks from a $D \x d$ matrix to a vector of length $d_k$, small enough to be generated per token from the vector stream, which we keep as the \emph{control path}.

\subsection{Architecture} \label{architecture}
LoomFormer keeps the backbone of the Transformer block and changes how sub-layers interface with the two streams: the vector stream carries control, i.e., where to attend and how to wire, and the matrix stream carries the content that is moved (\cref{fig:loom-arch}).
We describe the architecture bottom up, from ports to sub-layers to the model (pseudocode in \cref{pseudo code}), with $x$ denoting a sub-layer's pre-LN input.

\paragraph{MRead and MWrite.}
MRead and MWrite realize the matrix-stream ports of Eq.~(\ref{eq:port-duality}).
$\op{MRead}(M, x, x^r)$ generates a query $q^m \in \R^{d_k}$ to read $u \in \R^{d}$ from $M$, where $x^r = \op{GELU}(x W^R_1) \in \R^{D_r}$ with $W^R_1 \in \R^{D \x D_r}$ ($D_r \ll D$) is a low-rank feature shared by all MRead modules of a sub-layer to keep port generation cheap:
\begin{equation} \label{eq:MRead}
\begin{split}
    & q^m = \op{Norm}(x^r W^R_2) \odot \sigma(x W^R_g) + \bar q \odot \sigma(x W^R_{\bar g}) \\
    & u = q^m M
\end{split}
\end{equation}
Here $x^r$ is projected by $W^R_2 \in \R^{D_r \x d_k}$, normalized with RMSNorm and gated to get a dynamic query, which is summed with a static query $\bar q$ (also gated) to obtain $q^m \in \R^{d_k}$. The two scalar gates are computed from $x$ by projecting to scalars with $W^R_g, W^R_{\bar g} \in \R^{D \x 1}$ and then applying sigmoid.
Similarly, with a shared $x^w = \op{GELU}(x W^W_1) \in \R^{D_r}$, $\op{MWrite}(o, x, x^w)$ generates a key $k^m \in \R^{d_k}$ and writes a vector $o \in \R^{d}$ into $M$ as their outer product:
\begin{equation} \label{eq:MWrite}
\begin{split}
    & k^m = \op{Norm}(x^w W^W_2) \odot \sigma(x W^W_g) \\
    & \Delta M = k^m \otimes o
\end{split}
\end{equation}
Normalization and gating separate a port's direction from its magnitude: the normalized vector selects which slots to read or write, and the gate alone sets how much.

\paragraph{LoomMHA.}
To access the previous token's memory without horizontal attention, MShift first mixes $M_{[t]}$ with $M_{[t-1]}$ via a per-slot gate $z_{[t]} = \sigma(x_{[t]} W^S_g) \in \R^{d_k}$ ($W^S_g \in \R^{D \x d_k}$), and MNorm RMS-normalizes the matrix into $\widetilde{M}_{[t]}$.
Each head then reads four vectors from $\widetilde{M}$ with separate MRead modules: query, key, value and output bypass $o^m$:
\begin{equation} \label{eq:loom-read}
\begin{split}
    & \widetilde{M}_{[t]} = \op{MNorm}\big(z_{[t]} \odot M_{[t-1]} + (1 - z_{[t]}) \odot M_{[t]}\big) \\
    & \tilde{q}, \tilde{k}, v, o^m = \op{MRead}^{Q,K,V,O}(\widetilde{M}, x, x^r) \\
    & q, k = \op{RoPE}(x W^{Q,K}) \,\|\, (\tilde{q}, \tilde{k})
\end{split}
\end{equation}
where $W^Q, W^K \in \R^{D \x d/4}$ project $x$ to $d/4$ dimensions, to which rotary position embedding (RoPE) \cite{su2024roformer} is applied; the results are concatenated ($\|$) with $\tilde{q}$ and $\tilde{k}$ read from $M$.
The value is read from $M$ alone: the vector stream supplies only ports and RoPE queries/keys for slot and token routing, while all content moved by attention comes from $M$.
Unlike MHA, LoomMHA has two outputs: the projected head outputs for the vector stream, and the writes of all heads $i \in \{1, \dots, n\}$ summed into $\Delta M$:
\begin{equation} \label{eq:loom-attn}
\begin{split}
    & o_i = o^m_{i} + \op{Attention}(q_i, k_i, v_i) \\
    & \Delta M = \textstyle\sum_{i=1}^{n} \op{MWrite}_i(\op{Norm}(o_i), x, x^w) \\
    & \op{LoomMHA}(x, M) = \big(\op{Concat}_{i=1}^{n}[o_i] W^O,\, \Delta M\big)
\end{split}
\end{equation}
where $q_i, k_i, v_i, o^m_{i}$ are given by Eq.~(\ref{eq:loom-read}) for head $i$, and
$\op{Attention}$ is scaled dot-product attention over tokens with a strictly causal mask that excludes the current token: a token's own content reaches the head's output through the bypass $o^m_{i}$ instead.
LoomMHA is where vertical and horizontal attention are interwoven: the vertical reads from $M$ feed the horizontal attention of head $i$, and its output $o_i$ is written back to $M$ for later layers.

\paragraph{FFN with MWrite.}
The feed-forward network (FFN) sub-layer reads only $x$ and, like LoomMHA, returns a pair $(\op{FFN}_x(x), \Delta M)$, where $\op{FFN}_x$ is a standard FFN module.
A second, smaller $\op{FFN}_M$ computes the values to write, which are split into $n$ head-sized vectors ($\op{split}_i$) and written to $M$ with the sub-layer's MWrite modules.
This write is the only channel through which content of the vector stream enters $M$, since attention moves only content read from $M$.
\begin{equation} \label{eq:loom-mlp}
    \Delta M = \textstyle\sum_{i=1}^{n} \op{MWrite}_i(\op{split}_i(\op{FFN}_M(x)), x, x^w)
\end{equation}

\paragraph{LoomFormer.}
The embedding initializes both residual streams; $M_0$ is written from the embedding $x_0$ in the same way as the FFN writes:
\begin{equation} \label{eq:loom-embed}
\begin{split}
    & x_0 = \op{Embedding}(X) \\
    & M_0 = \textstyle\sum_{i=1}^{n} \op{MWrite}_i(\op{split}_i(x_0 W^E), x_0, x^w_0)
\end{split}
\end{equation}
where $W^E \in \R^{D \x D}$ is a value projection and $x^w_0 = \op{GELU}(x_0 W^W_1)$.
Each block $l$ then applies the two sub-layers with pre-LN, updating both streams residually (pairs are added element-wise):
\begin{equation} \label{eq:loom-block}
\begin{split}
    & (x'_l, M'_l) = (x_l, M_l) + \op{LoomMHA}(\op{LN}_1(x_l), M_l) \\
    & (x_{l+1}, M_{l+1}) = (x'_l, M'_l) + \op{FFN}(\op{LN}_2(x'_l))
\end{split}
\end{equation}
for $l = 0, \dots, L-1$.
Finally, $x_L$ is normalized and projected to produce the final output.
As $M_L$ is not used, the last block does not write to $M$.
We typically set $d_k = 2n$ slots (so $M$ has size $2D$) and $D_r = n d_k / 2 = n^2$ (a $2\X$ compression bottleneck relative to $n d_k$), matching the Transformer's parameter budget by adjusting FFN widths at an approximately $2{:}1$ ratio between $\op{FFN}_x$ and the $\op{FFN}_M$ write branch.

\subsection{Rewiring Capacity} \label{capacity}
On the matrix stream without MShift, reader $\mathsf{R}_j$ receives $u_j = q^m_{j} M = \sum_i \langle q^m_{j}, k^m_{i} \rangle\, o_i$ by Eq.~(\ref{eq:port-duality}).
The matrix stream thus keeps the division of roles of naive loom attention: its virtual weights are identities, and its wiring matrix is $C = K Q^\T$, of rank at most $d_k$, where the rows of $K$ and $Q$ are the keys $k^m_{i}$ of the writers and the queries $q^m_{j}$ of the readers.
\begin{proposition} \label{prop:ports}
(i) With ports fixed to $(K, Q)$, LoomFormer is a standard Transformer $\op{Tfm}_{K,Q}$ with residual stream $(x, \op{vec}(M)) \in \R^{D + d_k d}$, whose virtual weights absorb $K Q^\T$, hence node-gated.
(ii) With ports generated per token, LoomFormer is rewirable with one head per reader w.r.t.\ the wiring patterns with at most $d_k$ writers, whatever the total number of writers $w$.
\end{proposition}
In (ii), only the writers in the pattern are on, each taking a slot through a one-hot key while each reader's query selects the slots of its writers (\cref{app:ports}).
The $N(w, r-1)$ heads per reader that node gating needs (\cref{thm:cff}) are thus replaced by a per-token assignment of $d_k$ slots; when more than $d_k$ writers are on, slots are shared and writers interfere, as in linear attention.
By (i), this rewiring capacity comes from generating ports per token, not from the matrix memory itself.
Together, (i) and (ii) give a model-level view of LoomFormer's expressiveness: within each token, LoomFormer wires heads as $\op{Tfm}_{K(x), Q(x)}$, selecting per token one Transformer from the continuum $\{\op{Tfm}_{K,Q}\}$ much as a mixture of experts selects experts \cite{fedus2022switch}, except that the experts are wirings, share all other weights, and cost only $d_k$ numbers per port to select.

\begin{table*}[t]
\centering
\caption{LoomFormer among residual-stream mechanisms.
Ports $k^m$, $q^m$ are indexed by heads $\hdi$, $\hdj$ or, when shared across a layer, by layers $\hli$, $\hlj$ (and Q/K/V input $\hs$), and marked $(x)$ or $(M)$ when computed from the vector stream or from the memory itself; $\op{softmax}_{\hli}$ normalizes over writer layers.
Typical sizes assume $d_k = 4$ in HC and $d_k = 2n$ in LoomFormer (${+}D$ for its vector stream).}
\label{tab:design-space}
\begin{small}
\setlength{\tabcolsep}{4pt}
\begin{tabular}{lccc r@{${}={}$}l c}
\toprule
 & Effective wiring & Wiring rank & Granularity & \multicolumn{2}{c}{\makecell{Memory: slots $\x$ width\\$=$ typical size}} & \makecell{Residual\\stream} \\
\midrule
Transformer & $\bar C_{ij}$ & $1$ & -- & $1 \x D$ & $D$ & vector \\
DenseFormer & $\langle q^m_{\hlj}, k^m_{\hli} \rangle\, \bar C_{ij}$ & $1$ & layer & $l \x D$ & $lD$ & vector \\
HC / mHC & $\langle q^m_{\hlj}(M), k^m_{\hli}(M) \rangle\, \bar C_{ij}$ & $1$ & layer & $d_k \x D$ & $4D$ & matrix \\
AttnRes (full) & $\op{softmax}_{\hli} \langle q^m_{\hlj}, k^m_{\hli}(M) \rangle\, \bar C_{ij}$ & $1$ & layer & $2l \x D$ & $2lD$ & none \\
MUDDFormer & $\langle q^{m,\hs}_{\hlj}(x), k^m_{\hli} \rangle\, \bar C_{ij}$ & $3$ & layer $\x$ QKV & $l \x D$ & $lD$ & vector \\
\midrule
LoomFormer & $\langle q^{m,\hs}_{\hdj}(x), k^m_{\hdi}(x) \rangle\, I$ & $\min(d_k, 3n) = 2n$ & head $\x$ QKV & $d_k \x d$ & $2D \,({+}D)$ & matrix + vector \\
\bottomrule
\end{tabular}
\end{small}
\end{table*}

\section{Connections to Prior Work} \label{connections}
\cref{tab:design-space} compares LoomFormer with residual-stream mechanisms (derivations in \cref{app:designs}; extended related work in \cref{related_work}).
In DenseFormer \cite{pagliardini2024denseformer}, HC \cite{zhu2024hyper}, mHC \cite{xie2025mhc}, AttnRes \cite{kimi2026attention} and MUDDFormer \cite{xiao2025muddformer}, the ports are shared by all heads of a layer: apart from node gating and the fixed projections, their wiring depends on a writer only through its layer and on a reader only through its layer (and Q/K/V input in MUDDFormer).
Their wiring rank is therefore $1$, or $3$ in MUDDFormer, one per Q/K/V input, whereas LoomFormer generates its ports per head and per token and reaches $\min(d_k, 3n)$, i.e., $2n$ by default, with a smaller state than HC ($2D + D$ against $4D$).

\paragraph{Linear attention.}
Linear attention \cite{katharopoulos2020transformers,schlag2021linear} and its gated and delta-rule variants \cite{dao2024transformers,yang2024gated,yang2024parallelizing,yang2025gated,peng2024eagle} also read and write a fixed-size matrix state per head with keys and queries from the vector stream, but differ in axis and sharing.
Their state accumulates across tokens and its read is the head output, replacing horizontal attention, whereas $M$ accumulates along depth and its reads feed the query, key and value of horizontal attention.
Their state is also private to one head across tokens, so heads still compose through the static ports of the vector stream, whereas $M$ is shared by all heads of all layers at a token, as a channel through which heads read each other.
The two are thus orthogonal and can be combined.

\section{Experiments} \label{experiments}
We train all models on the Pile dataset \cite{gao2020pile} in JAX, initializing dynamic port gates near zero so training starts smoothly from static ports (\cref{hparams for lm}).
We evaluate scaling laws (\cref{scalinglaws}), large-scale pretraining and downstream performance (\cref{downstream}), mechanisms of the trained models (\cref{analyzing}), computational efficiency (\cref{overhead}) and ablations (\cref{ablation}).

\subsection{Scaling Laws} \label{scalinglaws}
\textbf{Settings.}
We train three model scales---Medium ($432\textrm{M}$), Large ($778\textrm{M}$) and XL ($1.4\textrm{B}$)---with sequence length $T = 4096$, batch size $0.5\textrm{M}$ tokens ($128$ sequences), and token budgets proportional to non-embedding parameters following \citet{hoffmann2022empirical} (architectural dimensions and optimizer hyperparameters in \cref{hparams for lm}).

\textbf{Baselines.}
We compare LoomFormer against recent residual-stream baselines on the same Transformer++ backbone \cite{touvron2023llama} (RoPE \cite{su2024roformer}, SwiGLU FFN \cite{shazeer2020glu}, RMSNorm, no linear bias):
DHC (dynamic HC with expansion rate $4$; \citealp{zhu2024hyper}), mHC \cite{xie2025mhc}, AttnRes (full) \cite{kimi2026attention}, and MUDDFormer \cite{xiao2025muddformer} (details in \cref{baselines for lm}).

\begin{figure}[tb]
\centering
\centerline{\includegraphics[width=0.9\columnwidth]{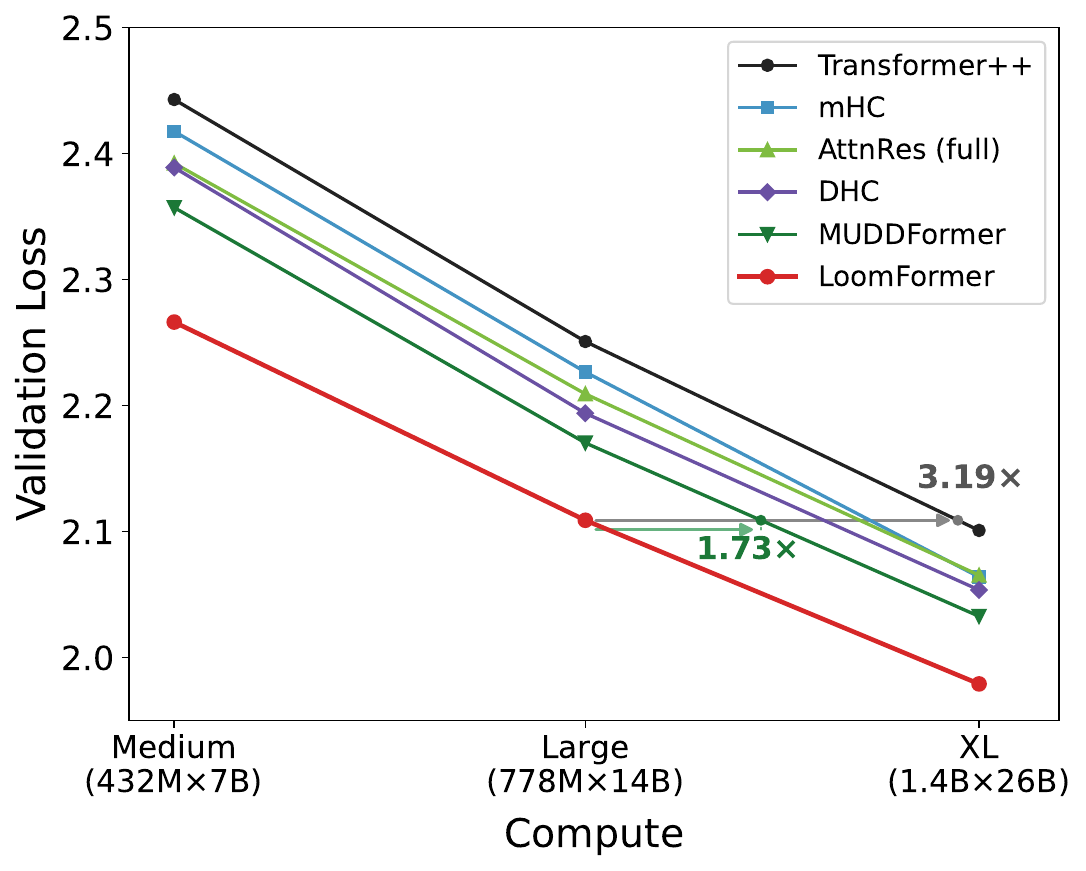}}
\vskip -0.1in
\caption{Scaling curves on Pile validation loss against training compute (total parameters $\x$ training tokens).} \label{fig:scaling-loomformer}
\vskip -0.1in
\end{figure}

\textbf{Results.}
Across the Medium, Large and XL scales (\cref{fig:scaling-loomformer}), mHC, AttnRes and DHC improve over Transformer++, and MUDDFormer improves further, while LoomFormer achieves the lowest loss at every scale.
Interpolating log-linearly along each baseline curve across Medium and Large, LoomFormer matches Transformer++ trained with 3.19--3.25\X\ compute (3.10\X\ net at Large after the 3.1\% FLOP overhead in \cref{tab:complexty analysis}) and all four residual-stream baselines at 1.73--2.76\X.
At XL, where LoomFormer lies below the largest model of every baseline, it lowers validation loss from $2.101$ to $1.979$ ($\Delta L = 0.122$, 1.8\X\ MUDDFormer's $0.068$) at 2.8\% extra FLOPs (\cref{tab:complexty analysis}).

\begingroup
\setlength{\tabcolsep}{2.0pt}
\begin{table*}[t]
\centering
\caption{Zero/five-shot downstream evaluation results ($\Delta$acc is over Pythia-2.8B; best in bold within each block).}
\label{tab:downstream}
\begin{small}
\begin{tabular}{l
>{\raggedright}p{0.7cm}>{\raggedright}p{0.85cm}|c
>{\centering\arraybackslash}p{0.9cm}>
{\centering\arraybackslash}p{0.73cm}>{\centering\arraybackslash}p{1cm}>{\centering\arraybackslash}p{0.75cm}>{\centering\arraybackslash}p{0.75cm}>{\centering\arraybackslash}p{0.64cm}>{\centering\arraybackslash}p{0.79cm}>{\centering\arraybackslash}p{0.85cm}>{\centering\arraybackslash}p{0.85cm}>{\centering\arraybackslash}p{0.85cm}l}
\toprule
Model & \makecell{Pile \\ ppl$\downarrow$} & \makecell{FLAN \\ ppl$\downarrow$} & & \makecell{LAM \\ BADA} & PIQA & \makecell{Wino \\ Grande} & \makecell{ARC \\ -E} & \makecell{ARC \\ -C} & SciQ & BoolQ & \makecell{Hella \\ Swag} & \makecell{RACE \\ -M} & \makecell{RACE \\ -H} & \makecell{Avg \\ acc$\uparrow$ /$\Delta$acc} \\
\midrule
Pythia-2.8B          & 6.63 & 8.16 & \multirow{5}{*}{\textit{0-shot}} & 64.7 & 73.9 & 59.4 & 64.4 & 29.5 & 88.2 & 64.5 & 45.4 & 38.1 & 34.9 & 56.3\\
Pythia-6.9B          & 6.29 & 7.85 & & 67.3 & 75.2 & 60.9 & 67.3 & 31.3 & 89.7 & 63.7 & 48.0 & 40.6 & 37.0 & 58.1\\
mHC-Pythia-2.8B      & 6.24 & 7.35 & & 68.5 & 74.4 & 62.8 & 67.1 & 30.7 & 90.5 & 66.9 & 47.5 & \B{42.3} & 37.0 & 58.8/+2.5\\
Pythia-12B           & \B{6.01} & 7.26 & & 70.5 & \B{76.0} & \B{63.9} & \B{70.2} & \B{31.8} & 90.2 & \B{67.4} & \B{50.3} & 40.6 & 38.3 & \B{59.9}\\
\B{LoomPythia-2.8B}  & 6.02 & \B{6.94} & & \B{71.8} & 74.7 & 63.7 & 68.5 & 31.7 & \B{92.4} & 65.8 & 49.1 & \B{42.3} & \B{39.0} & \B{59.9}/+3.6\\
\midrule
Pythia-2.8B          & - & - & \multirow{5}{*}{\textit{5-shot}} & 60.5 & 73.6 & 60.6 & 67.3 & 32.3 & 94.3 & 65.6 & 45.1 & 38.4 & 35.6 & 57.3\\
Pythia-6.9B          & - & - & & 63.8 & 75.5 & 63.7 & 70.2 & 35.6 & 95.1 & 65.7 & 48.1 & 39.0 & 36.5 & 59.3\\
mHC-Pythia-2.8B      & - & - & & 62.4 & 74.9 & 64.1 & 70.8 & 35.5 & 95.6 & 69.4 & 47.6 & 40.3 & 38.2 & 59.9/+2.6\\
Pythia-12B           & - & - & & 67.3 & \B{76.0} & 64.2 & \B{71.0} & \B{36.5} & 95.3 & 68.0 & \B{50.3} & 40.1 & 38.8 & 60.8\\
\B{LoomPythia-2.8B}  & - & - & & \B{69.0} & 75.7 & \B{65.7} & 70.8 & 35.1 & \B{96.0} & \B{71.2} & 49.5 & \B{43.1} & \B{40.1} & \B{61.6}/+4.3\\
\bottomrule
\end{tabular}
\end{small}
\vskip -0.08in
\end{table*}
\endgroup

\begin{figure*}[!b]
\centering
\includegraphics[width=\textwidth]{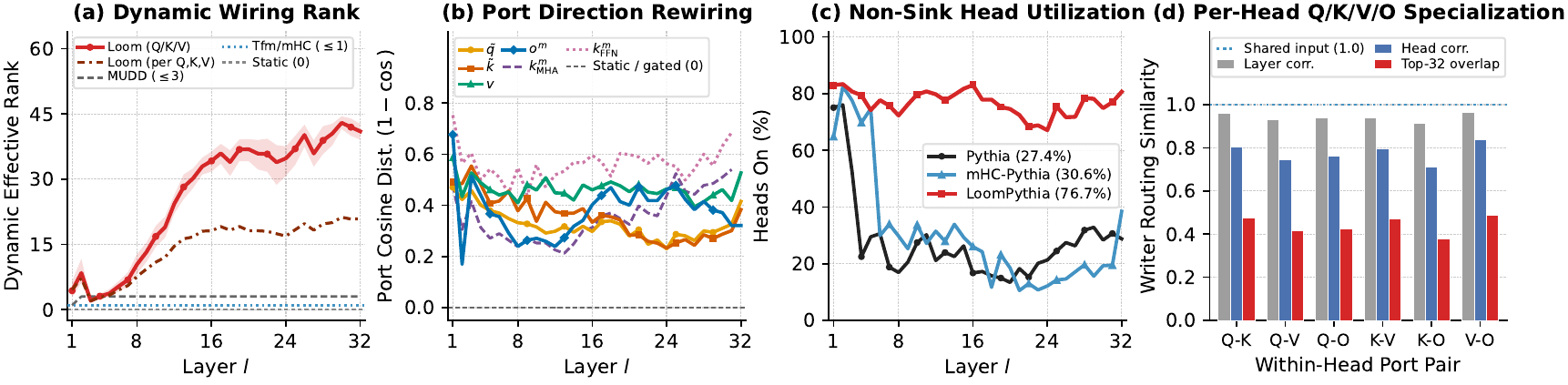}
\vskip -0.08in
\caption{Mechanistic comparison of 2.8B models across layers $l = 1, \dots, 32$ on Pile validation sequences (definitions in \cref{interpretation}).
(a) Energy effective rank of the token-centered dynamic Q/K/V wiring matrix (band: 10th--90th percentiles) and per single Q, K or V input, vs.\ the bounds of \cref{tab:design-space}.
(b) Cross-token cosine distance ($1 - \cos$) of read ports ($\tilde{q}, \tilde{k}, v, o^m$) and write ports ($k^m_{\mathrm{MHA}}, k^m_{\mathrm{FFN}}$) in $\R^{d_k}$ (static/gated $= 0$).
(c) Percentage of heads that are on (non-sink; legend: 32-layer mean).
(d) Within-head writer routing similarity across port pairs in layers 3--32.}
\label{fig:mech-main}
\vskip -0.06in
\end{figure*}

\subsection{Large-Scale Pretraining and Evaluation} \label{downstream}

\textbf{Settings.}
To test whether LoomFormer's gains persist under extended training, we compare against the Pythia suite \cite{biderman2023pythia} on 300B Pile tokens (over 5\X\ the Chinchilla budget of \cref{scalinglaws}; \citealp{hoffmann2022empirical}).
Specifically, we train LoomPythia-2.8B and mHC-Pythia-2.8B---the residual-stream baseline that also keeps a depth-independent state size (\cref{tab:design-space})---and compare them with Pythia 2.8B--12B.
Except for using serial instead of parallel sub-layers\footnote{mHC requires serial sub-layers by design, and parallel vs.\ serial FFN placement is quality-neutral for standard Transformers, as shown by \citet{chowdhery2023palm} and our own experiments.}, both LoomPythia and mHC-Pythia use the same architecture choices and training hyperparameters as Pythia \citep[Appendix~E]{biderman2023pythia}.
We evaluate zero- and five-shot accuracy across 10 standard benchmarks.

\textbf{Results.}
As shown in \cref{tab:downstream} and \cref{fig:loompythia-downstream}, LoomPythia-2.8B improves average accuracy over Pythia-2.8B by $+3.6$ in 0-shot and $+4.3$ in 5-shot ($+3.95$ on their mean), achieving 1.6\X\ the gain of mHC-Pythia-2.8B ($+2.5$ / $+2.6$, mean $+2.55$).
Its 0-shot average ($59.9$) matches Pythia-12B ($59.9$), its 5-shot average ($61.6$) exceeds Pythia-12B ($60.8$), and its 0/5-shot mean ($60.75$) matches a log-linear extrapolation of Pythia at \tildemid4.9\X\ compute.
On pretraining metrics, LoomPythia-2.8B reaches Pile ppl $6.02$, on par with Pythia-12B ($6.01$) and well ahead of mHC-Pythia-2.8B ($6.24$).

As an out-of-domain evaluation, \cref{tab:downstream} also reports target-span perplexities on a randomly sampled subset of the FLAN Collection \cite{longpre2023flan}, which features instruction following, chain-of-thought and in-context few-shot data.
The advantage of LoomPythia on FLAN is even larger than on Pile: FLAN ppl $6.94$ is best in the table, beating both Pythia-12B ($7.26$) and mHC-Pythia-2.8B ($7.35$).
Whereas mHC's gain barely changes from 0-shot to 5-shot ($+2.5$ vs.\ $+2.6$), LoomPythia's widens from $+3.6$ to $+4.3$, indicating stronger in-context learning that \cref{analyzing} traces to per-token rewiring freeing deep heads from attention sinks.

\subsection{Mechanistic Analysis} \label{analyzing}

We analyze Pythia-2.8B, mHC-Pythia-2.8B and LoomPythia-2.8B (\cref{downstream}) on held-out Pile validation sequences; metric definitions and additional visualizations are in \cref{interpretation}.

\textbf{Per-token circuit rewiring.}
After subtracting each pair's token-mean static component, the dynamic wiring matrix on the $3n = 96$ Q/K/V readers of layer $l$ has rank $0$ under fixed ports (\cref{prop:ports}(i)), at most $1$ in Transformer and mHC, at most $3$ in MUDDFormer, and up to $\min(d_k, 3n) = 64$ in LoomFormer (\cref{tab:design-space}).
In \cref{fig:mech-main}(a), LoomPythia-2.8B's dynamic effective rank rises with depth to $41.0$ ($26.4$ mean), placing $71.0\%$ of dynamic wiring energy ($90.7\%$ at layer 32) outside rank $1$, and averages $14.3$ ($20.9$ at layer 32) within a single Q, K or V input where MUDDFormer also has rank $\le 1$.
\cref{fig:mech-main}(b) directly tests \cref{prop:ports}(i) vs.\ (ii) in $\R^{d_k}$: whereas static or scalar-gated ports have zero cross-token cosine distance, LoomPythia's read ports $\tilde{q}, \tilde{k}, v, o^m$ average $0.33$--$0.46$ (up to $0.68$) and its attention and $\op{FFN}_M$ write ports average $0.36$ and $0.56$ (up to $0.75$; \cref{interpretation} traces an in-context circuit rewiring across adjacent tokens).

\textbf{Head utilization and within-head specialization.}
In a node-gated model (\cref{framework}, \cref{thm:cff}), heads whose fixed inputs are not needed at a token turn themselves off by attending to sink tokens \cite{vig2019analyzing,xiao2023efficient}.
Counting non-sink heads as on \cite{xiao2025muddformer}, \cref{fig:mech-main}(c) shows that in layers 17--32, only $22.9\%$ of heads are on in Pythia-2.8B and $17.7\%$ in mHC-Pythia-2.8B---whose four streams still merge into one shared input per layer---whereas LoomPythia-2.8B keeps $74.4\%$ on ($3.3\X$ and $4.2\X$), more than doubling content-token attention mass and accounting for its 5-shot and FLAN gains (\cref{downstream}).
Finally, \cref{fig:mech-main}(d) verifies $\text{head} \x \text{QKV}$ specialization (\cref{tab:design-space}): although routing looks correlated at the layer level ($0.92$--$0.96$), per-writer correlation within the same head drops to $0.71$--$0.84$ and top-32 writer overlap is $0.38$--$0.49$, so over half of the top writers read by a head's $\tilde{q}, \tilde{k}, v, o^m$ are distinct.

\begin{table}[t]
\centering
\caption{Analytical computational overhead ($+\textrm{FLOPs}$) and measured relative throughput of LoomFormer vs.\ matched-size Transformer at sequence length $T$ (absolute speeds in \cref{mem_usage}).}
\label{tab:complexty analysis}\label{tab:overheadl}
\begin{small}
\setlength{\tabcolsep}{4.1pt}
\begin{tabular}{@{}lcccc@{}}
\toprule
Model & \makecell{$+\textrm{FLOPs}$\\(\%)} & \makecell{Train (\%)\\v5p} & \makecell{Prefill (\%)\\bs $1$\,/\,$8$} & \makecell{Decode (\%)\\bs $1$\,/\,$16$\,/\,$64$} \\
\midrule
Medium & $3.70$ & $90.5$ & $87.4$\,/\,$90.4$ & $91.7$\,/\,$89.2$\,/\,$92.3$ \\
Large  & $3.13$ & $89.2$ & $85.3$\,/\,$86.2$ & $85.4$\,/\,$87.8$\,/\,$84.9$ \\
XL     & $2.77$ & $92.3$ & $83.8$\,/\,$85.3$ & $86.9$\,/\,$84.6$\,/\,$86.6$ \\
2.8B   & $1.95$ & $92.4$ & $87.5$\,/\,$90.5$ & $86.4$\,/\,$87.6$\,/\,$88.9$ \\
\bottomrule
\end{tabular}
\end{small}
\vskip -0.1in
\end{table}

\subsection{Computational Complexity and Efficiency} \label{overhead}

\textbf{Analytical computational overhead.}
Under a matched parameter budget, LoomFormer adds \tildemid$\frac{1}{d(1 + \rho / 12)} + \frac{1}{8(1 + 12 / \rho)}$ relative forward FLOPs from reading/writing $M$ and wider Q/K ($d_k = 2n$, $\rho = T / D$; derived in \cref{append:complexity analysis}), totaling $1.95\%$--$3.70\%$ across our four scales (\cref{tab:complexty analysis}).

\textbf{Wall-clock training and inference throughput.}
\cref{tab:complexty analysis} reports relative throughput for JAX training on TPU v5p pods (with fused Pallas kernels) and BF16 PyTorch inference on an NVIDIA A800 GPU (with a fused Triton kernel and \texttt{torch.compile}; \cref{mem_usage}).
LoomFormer sustains $89\%$--$92\%$ of Transformer training speed and $84\%$--$92\%$ of inference speed across prefill and decode batch sizes; the gap beyond the $2.0\%$--$3.7\%$ FLOPs reflects HBM I/O on $M$, lower arithmetic intensity of per-token $M$ contractions, and smaller split projections and norms, and is far outweighed by the over \mbox{3\X} compute savings (\cref{scalinglaws,downstream}).

\subsection{Ablation Studies} \label{ablation}
We ablate LoomFormer at the Medium scale (432M, 7B tokens; \cref{scalinglaws}), reallocating saved parameters to $\op{FFN}_x$ to match the parameter budget, and report Pile validation loss and the retained gain $\textrm{Gain} = (L_{\textrm{Tfm++}} - L) / (L_{\textrm{Tfm++}} - L_{\textrm{Loom}})$ (\cref{tab:ablation}; full suite in \cref{app:ablations}).

\begin{table}[t]
\centering
\caption{Ablations of LoomFormer at the Medium scale.}
\label{tab:ablation}
\begin{small}
\setlength{\tabcolsep}{3.5pt}
\begin{tabular}{lccc}
\toprule
Variant & Loss$\downarrow$ & $\Delta$Loss & Gain \\
\midrule
\B{LoomFormer} (full) & \B{2.266} & --- & \B{100.0\%} \\
Transformer++ (baseline) & 2.443 & $+$0.177 & 0.0\% \\
\midrule
\multicolumn{4}{l}{\textit{Dynamic rewiring (\cref{prop:ports})}} \\
\hspace{4pt}Static write keys ($k^m$) & 2.283 & $+$0.017 & 90.3\% \\
\hspace{4pt}Static read queries ($q^m$) & 2.310 & $+$0.043 & 75.4\% \\
\hspace{4pt}Static read \& write ($q^m, k^m$) & 2.344 & $+$0.078 & 56.0\% \\
\midrule
\multicolumn{4}{l}{\textit{Port normalization \& gating (Eqs.~\ref{eq:MRead}--\ref{eq:MWrite})}} \\
\hspace{4pt}$-$ dynamic port gates & 2.277 & $+$0.011 & 93.7\% \\
\hspace{4pt}$-$ dynamic port $\op{Norm}$ & 2.289 & $+$0.023 & 87.1\% \\
\hspace{4pt}$-$ dynamic port gates \& $\op{Norm}$ & 2.283 & $+$0.017 & 90.3\% \\
\midrule
\multicolumn{4}{l}{\textit{Sub-layer components (\cref{architecture})}} \\
\hspace{4pt}$-$ $\op{FFN}_M$ write & 2.300 & $+$0.034 & 80.7\% \\
\hspace{4pt}$-$ output bypass $o^m$ & 2.299 & $+$0.032 & 81.8\% \\
\hspace{4pt}$-$ vector-stream RoPE $W^{Q,K}$ & 2.286 & $+$0.020 & 88.7\% \\
\hspace{4pt}$-$ $\op{MShift}$ & 2.277 & $+$0.010 & 94.2\% \\
\bottomrule
\end{tabular}
\end{small}
\vskip -0.1in
\end{table}

\textbf{Dynamic rewiring.}
Replacing dynamic $k^m$, dynamic $q^m$, or both with learned static vectors (dropping their gates) tests \cref{prop:ports}.
Fixing both reduces LoomFormer to $\op{Tfm}_{K,Q}$ and cuts the gain to 56.0\%---still ahead of layer-wise dynamic baselines (\cref{fig:scaling-loomformer})---so the static head-level dual-stream backbone and per-token rewiring contribute roughly equal halves.
Fixing $q^m$ hurts far more than fixing $k^m$ (75.4\% vs.\ 90.3\% retained): writers can use stable slots if readers select among them per token.

\textbf{Port normalization and gating.}
Removing either the dynamic-branch gates or port RMSNorm hurts, yet removing both beats removing RMSNorm alone (90.3\% vs.\ 87.1\%): an unnormalized projection already varies in norm, making a separate gate redundant.
Pairing RMSNorm with gating works best by separating slot direction from read/write magnitude.

\begin{figure}[!b]
\centering
\includegraphics[width=0.92\columnwidth]{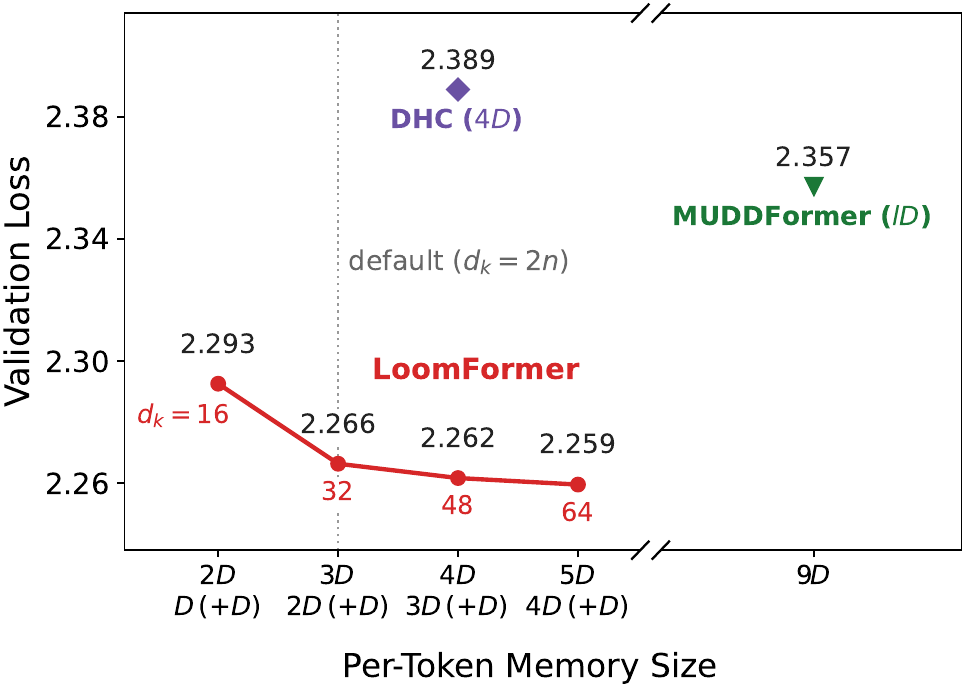}
\vskip -0.08in
\caption{Validation loss vs.\ per-token memory size and matrix slots $d_k$ at the Medium scale ($L = 18$).}
\label{fig:m-capacity}
\vskip -0.1in
\end{figure}

\textbf{Sub-layer components.}
Removing $\op{FFN}_M$ (Eq.~\ref{eq:loom-mlp}) hurts most, as the sole post-embedding path from the vector stream into $M$, followed closely by removing the bypass $o^m$ and restoring self-attention to the current token (Eq.~\ref{eq:loom-attn}).
Applying RoPE to $d/4$ of the $M$-read coordinates instead of $x W^{Q,K}$ (Eq.~\ref{eq:loom-read}) also hurts, supporting separate streams for positions and content, whereas removing $\op{MShift}$ still retains 94.2\% of the gain.

\textbf{Memory capacity.}
As we vary the slot count $d_k$ (\cref{fig:m-capacity}), validation loss drops sharply ($-0.026$) from $d_k = 16$ to $32$ with diminishing returns beyond, placing our default $d_k = 2n$ at the knee given the linear growth of MRead/MWrite FLOPs (\cref{append:complexity analysis}) and I/O with $d_k$.
Even at $d_k = 16$ ($2D$ total state), LoomFormer already outperforms DHC ($4D$) and MUDDFormer ($lD$, averaging $9D$ across $L = 18$ layers), showing that head-level rewiring uses state memory far more effectively than layer-wise mechanisms.

\section{Conclusion}
Viewing Transformer circuits through their dynamic wiring matrix shows that standard Transformers and layer-wise residual-stream mechanisms are restricted to low-rank gating, requiring many heads per reader to avoid circuit interference.
Loom attention lifts this bottleneck by interweaving vertical linear attention over a per-token matrix memory with horizontal attention over tokens, making circuits rewirable per token at one head per reader and $2.0\%$--$3.7\%$ extra forward FLOPs.
Across model scales and 300B-token pretraining, LoomFormer matches Transformers trained with over \mbox{3\X} compute and recent residual-stream baselines with \mbox{\tildemid1.7--2.8\X}, while keeping over three quarters of its attention heads on and rewiring its ports from token to token.
Because the matrix memory operates along depth at each token, combining loom attention with linear or sparse attention along the sequence is a natural direction for future foundation models.

\bibliography{main}
\bibliographystyle{icml2027}

\onecolumn
\appendix
\section{Theory of Rewirable Circuits} \label{app:theory}
\subsection{Normalization and proof of Proposition~\ref{prop:rectangle}} \label{app:node-gated}
\begin{lemma} \label{lemma:fixed}
In a Transformer with pre-LN (LayerNorm or RMSNorm), the contribution of earlier writers to reader $\mathsf{R}_j$'s input $u_j = \op{LN}(x) W^s_j$ is $\lambda(x) \sum_i o_i \bar C_{ij}$, where the scale $\lambda(x) > 0$ is shared by all writers and the virtual weights $\bar C_{ij}$ are fixed.
\end{lemma}
\begin{proof}
Write $x = \sum_i o_i W^O_i + x_{\mathrm{other}}$, where $x_{\mathrm{other}}$ sums the embedding and earlier FFN outputs.
LayerNorm computes $\op{LN}(x) = \lambda(x)\,(x - \mu(x)\mathbf{1}) \odot \gamma + \beta$ with $\lambda(x) = \sqrt{D} / \|x - \mu(x)\mathbf{1}\| > 0$ (putting $\lambda(x) = 0$ on the degenerate set $x \in \op{span}(\mathbf{1})$).
Since centering is the fixed linear map $x - \mu(x)\mathbf{1} = x \big(I - \frac{1}{D}\mathbf{1}\mathbf{1}^\T\big)$,
\[
    u_j = \op{LN}(x) W^s_j
    = \lambda(x) \textstyle\sum_i o_i \bar C_{ij} + \lambda(x)\, x_{\mathrm{other}} \big(I - \tfrac{1}{D}\mathbf{1}\mathbf{1}^\T\big) \op{diag}(\gamma) W^s_j + \beta W^s_j ,
\]
with fixed virtual weights $\bar C_{ij} = W^O_i \big(I - \frac{1}{D}\mathbf{1}\mathbf{1}^\T\big) \op{diag}(\gamma) W^s_j$ absorbing the linear maps of the normalization (any projection bias adds to $\beta W^s_j$).
For RMSNorm, centering is omitted, $\beta = 0$ and $\lambda(x) = \sqrt{D} / \|x\|$.
\end{proof}
\begin{proof}[Proof of Proposition~\ref{prop:rectangle}]
By \cref{lemma:fixed}, pre-LN rescales all writers of a token by the same $\lambda(x) > 0$.
Within a token, what depends on the input is only this scale and which heads are on: let $g_i = \lambda(x)$ if the head of $\mathsf{W}_i$ is on and $0$ otherwise, and $g'_j = 1$ if the head of $\mathsf{R}_j$ is on and $0$ otherwise.
Then $C = g\, g'^\T$, which is Eq.~(\ref{eq:node-gated}), and since $\lambda(x) > 0$, $\mathsf{W}_i$ and $\mathsf{R}_j$ are connected ($g_i g'_j \bar C_{ij} \ne 0$) if and only if both are on and $\bar C_{ij} \ne 0$.
\end{proof}

\subsection{Full statement and proof of Theorem~\ref{thm:cff}} \label{app:cff}
\paragraph{Setting.}
The role of reader $\mathsf{R}_j$ ($j \in [r]$, $r \ge 2$) is played by a set $H_j$ of heads, so a reader is here a role rather than one input of one head; by Eq.~(\ref{eq:node-gated}), head $h \in H_j$ \emph{receives} $\mathsf{W}_i$ when its effective wiring $g_i g'_h \bar C_{ih}$ is nonzero.
A wiring task assigns to each reader a writer $\pi(j) \in [w]$, and the tasks are all maps $\pi: [r] \to [w]$.
Reader $\mathsf{R}_j$ \emph{receives exactly} $\mathsf{W}_{\pi(j)}$ if some $h \in H_j$ receives $\mathsf{W}_{\pi(j)}$ and no $h \in H_j$ receives any $\mathsf{W}_i$ with $i \ne \pi(j)$.
We require every head in $H_j$ that is on to be free of other writers because query and key inputs feed a softmax, so interference at one head cannot be cancelled downstream.
Every writer in $\pi([r])$ must therefore be on ($g_{\pi(j)} \ne 0$), and turning others on can only add interference, so we may assume the writers that are on are $\pi([r])$.
For a fixed reader $\mathsf{R}_j$, let $B_i = \{h \in H_j : \bar C_{ih} \ne 0\}$.
A family $\{B_i\}_{i=1}^{w}$ is $t$-cover-free if $B_i \not\subseteq \bigcup_{i' \in J} B_{i'}$ for every $i \in [w]$ and every $J \subseteq [w] \setminus \{i\}$ with $|J| \le t$.
\begin{theorem}[full version of \cref{thm:cff}] \label{thm:cff-full}
For $r \ge 2$, a node-gated architecture with $w$ writers realizes all wiring tasks $\pi: [r] \to [w]$ if and only if, for every reader, $\{B_i\}_{i=1}^{w}$ is an $(r-1)$-cover-free family.
Hence each reader needs at least $N(w, r-1)$ heads, and this number suffices.
\end{theorem}
\begin{proof}
\emph{If.} Given $\pi$, turn on only the writers $\pi([r])$.
For reader $\mathsf{R}_j$, let $J = \pi([r]) \setminus \{\pi(j)\}$, so $|J| \le r - 1$.
Since $\{B_i\}$ is $(r-1)$-cover-free, there is a head $h \in B_{\pi(j)} \setminus \bigcup_{i \in J} B_i$; turn on $h$ and no other head of $H_j$.
Then $h$ receives $\mathsf{W}_{\pi(j)}$ and none of the other writers that are on.

\emph{Only if.} Fix a reader $\mathsf{R}_j$, a writer $i$ and a set $J \subseteq [w] \setminus \{i\}$ with $|J| \le r - 1$.
Choose $\pi$ with $\pi(j) = i$ that maps the other $r - 1$ readers onto $J$, which is possible since $r - 1 \ge 1$, $|J| \le r - 1$ and different readers may share a writer (if $J = \emptyset$, map them to $i$).
Realizing $\pi$ requires a head $h \in H_j$ that is on with $\bar C_{ih} \ne 0$ and $\bar C_{i'h} = 0$ for all $i' \in J$, i.e., $h \in B_i \setminus \bigcup_{i' \in J} B_{i'}$.
Hence $B_i \not\subseteq \bigcup_{i' \in J} B_{i'}$, and the family is $(r-1)$-cover-free.

\emph{Counting.} The sets $B_i$ are subsets of $H_j$, so $|H_j| \ge N(w, r-1)$.
Conversely, given an $(r-1)$-cover-free family of $w$ sets on a ground set of size $N(w, r-1)$ for each reader, create one head per point and set $\bar C_{ih} = I_d$ if $h \in B_i$ and $0$ otherwise; this is realized in $D = \min\{w, r N(w, r-1)\}\, d$ dimensions by assigning orthogonal $d$-dimensional subspaces either to the $w$ writers or to the $r N(w, r-1)$ heads.
\end{proof}
\paragraph{Remarks.}
(i) Attention sinks gate heads off only approximately; under the threshold of \cref{framework}, the counting argument is unchanged.
(ii) Sufficiency at $N(w, r-1)$ heads per reader holds in a Transformer when each of the $r$ readers belongs to a separate head (with its other two inputs fixed); when $r = 3n$ counts all Q, K and V inputs of $n$ heads, coupling Q, K and V inside the same physical head (and sink gate $g'_h$) can only require more heads, while the lower bound $N(w, r-1)$ still applies.
Likewise, if several heads of a value reader share identical attention weights, their outputs add linearly and interference could in principle cancel, whereas our clean-head requirement is necessary for query and key inputs.
(iii) The model excludes routing through FFNs, discussed in \cref{app:ffn-routing}.

\subsection{Bounds on $N(w, t)$ and remarks on Theorem~\ref{thm:cff}} \label{app:bounds}
\paragraph{Bounds.}
A $1$-cover-free family is an antichain, so by Sperner's theorem \cite{sperner1928satz} $N(w, 1)$ is the smallest $N$ with $\binom{N}{\lfloor N/2 \rfloor} \ge w$, about $\log_2 w$.
For $2 \le t \le \sqrt{w}$, $N(w, t) = \Omega(t^2 \log w / \log t)$ \cite{dyachkov1982bounds,ruszinko1994upper,furedi1996r} and $N(w, t) = \mathcal{O}(\min\{t^2 \log w, w\})$ \cite{erdos1985families}; for $t > \sqrt{w}$, Bassalygo's bound $N(w, t) \ge \min\{\binom{t+2}{2}, w\}$ \cite{dyachkov1982bounds} together with the singleton construction $N(w, t) \le w$ gives $N(w, t) = \Theta(w)$, reaching $N(w, t) = w$ (one head per writer) once $\binom{t+2}{2} \ge w$.
\paragraph{Group testing.}
\cref{thm:cff-full} is an instance of non-adaptive group testing: the heads of a reader are tests that must isolate the selected writer from up to $r-1$ other writers that are on.
\paragraph{Dense and sparse layouts.}
With a single head per reader ($|H_j| = 1$), the two layouts of \cref{fig:rewire} are degenerate codes: $B_1 = \dots = B_w = H_j$ (dense), which is not $1$-cover-free and makes circuits interfere, and $B_i = \emptyset$ for some writer $i$ (sparse), which leaves $(i, j)$ unreachable.
\paragraph{Superposition.}
At layer scale ($r = 3n$, $N(w, r-1) = \Theta(w)$), separating $w$ writers as above takes $\Theta(wd)$ dimensions ($wd$ with orthogonal writer subspaces).
With $nL$ writers sharing $D = nd$ dimensions, writers across layers must overlap in superposition \cite{elhage2022toy}, so virtual weights are rarely exactly zero and in practice part of the price is paid as interference.

\subsection{Cost of naive loom attention} \label{app:cost}
Indexing layers by $l = 0, \dots, L - 1$ as in \cref{architecture}, every reader of layer $l$ attends over $nl$ earlier head writers.
Over the $3n$ readers of all $L$ layers, the projections $W^\alpha$ of Eq.~(\ref{eq:vertical}) take $\sum_{l=0}^{L-1} 3n \cdot nl \cdot D = \frac{3}{2} n^2 L(L-1) D \approx \frac{3}{2} n^2 L^2 D$ parameters, and as many multiply-adds per token.
This grows quadratically in the total number of heads $nL$ and rivals the $12 L D^2$ of the whole model once $n^2 L \approx 8D$ (i.e., $nL \approx 8d$), e.g., $70\%$ of it for our XL configuration ($L = 28$, $n = 20$, $D = 1920$).
Factorizing the weights into keys and queries removes $W^\alpha$, but still leaves $\approx \frac{3}{2} n^2 L^2$ attention weights per token to compute and store for backpropagation, plus $\approx \frac{3}{2} n L^2 D$ multiply-adds to aggregate the $o_i$'s.

\subsection{Routing through FFNs} \label{app:ffn-routing}
An intermediate FFN between the writers and the reader layer can route any pattern $P \subseteq [w] \x [r]$ by copying $o_i$ into a $d$-dimensional subspace private to $\mathsf{R}_j$ whenever $(i, j) \in P$.
Concretely, suppose the residual stream $x \in \R^D$ carries binary indicators $p_{ij} \in \{0, 1\}$ of $P$ ($p_{ij} = 1$ iff $(i, j) \in P$) and the writer outputs satisfy $\|o_i\|_\infty \le \Lambda$ (ignoring pre-LN rescaling).
Since $\op{ReLU}\big(z - \Lambda(1 - p)\big) - \op{ReLU}\big(-z - \Lambda(1 - p)\big)$ equals $z$ when $p = 1$ and $0$ when $p = 0$ for every $z \in [-\Lambda, \Lambda]$, a two-layer ReLU FFN can compute the gated copy $p_{ij} o_i \in \R^d$ coordinate-wise for each pair $(i, j)$ and sum over $i$ into $\mathsf{R}_j$'s private subspace.
However, this construction uses $2 w r d$ hidden units---hence $4 w r d D = \mathcal{O}(w r d D)$ parameters, or $12 n l D^2$ when routing $w = nl$ writers to $r = 3n$ readers ($n$ times the $12 l D^2$ parameters of all $l$ preceding blocks)---requires an intervening FFN step, and needs disjoint subspaces for the $w$ writers, $r$ readers and $wr$ control coordinates ($D \ge (w + r)d + wr$).
We give this construction only as an upper bound and do not prove a lower bound on FFN routing.

\subsection{Proof of Proposition~\ref{prop:ports}} \label{app:ports}
\paragraph{Part (i).}
Flatten $M \in \R^{d_k \x d}$ row by row into $\op{vec}(M) \in \R^{d_k d}$, so that entry $(a, b)$ of $M$ is entry $(a-1)d + b$ of $\op{vec}(M)$.
For a static read port $q \in \R^{d_k}$, $(qM)_b = \sum_a q_a M_{ab} = \big(\op{vec}(M)\,(q^\T \otimes I_d)\big)_b$.
For a static write port $k \in \R^{d_k}$, $\op{vec}(k^\T o)_{(a-1)d+b} = k_a o_b = \big(o\,(k \otimes I_d)\big)_{(a-1)d+b}$.
Hence every read and write of $M$ is a fixed linear map between $\op{vec}(M)$ and the head, i.e., a Transformer projection with Kronecker structure, $q^\T \otimes I_d$ for reads and $k \otimes I_d$ for writes, and the composition weight between $\mathsf{W}_i$ and $\mathsf{R}_j$ is the constant $\langle q^m_{j}, k^m_{i} \rangle$.
The coupling $K Q^\T$ is therefore fixed and belongs to the virtual weights, $\bar C_{ij} = \langle q^m_{j}, k^m_{i} \rangle I_d$; only gating heads through attention sinks (together with normalization scales) depends on the input, as in Eq.~(\ref{eq:node-gated}), so $\op{Tfm}_{K,Q}$ is node-gated.
In the full architecture (\cref{architecture}), $\op{MNorm}$ is RMSNorm over $\op{vec}(M)$ and rescales all matrix-stream writers by one shared factor (\cref{lemma:fixed}), while the bypass $o^m_i$ and $(\op{FFN}_x, \op{FFN}_M)$ become a fixed linear skip and a standard FFN on $(x, \op{vec}(M))$.
The per-head $\op{Norm}(o_i)$ on LoomMHA writes (Eq.~\ref{eq:loom-attn}) applies a fixed diagonal gain (absorbed into the write projection) and rescales writer $\mathsf{W}_i$ by $\sqrt{d}/\|o_i\|$, and even if the scalar port gates $\sigma(x W^W_g), \sigma(x W^R_{\bar g})$ (Eqs.~\ref{eq:MRead}--\ref{eq:MWrite}) are retained on static port directions, they multiply whole rows or columns of $C$, leaving $C = g\, g'^\T$ of rank at most one (cf.\ \cref{app:ablations}).
With ports generated per token, the computation at each token and layer coincides with that of $\op{Tfm}_{K(x), Q(x)}$.
The coincidence holds within a token; across tokens, attention mixes values computed by different members, so the model as a whole is not any single $\op{Tfm}_{K,Q}$.

\paragraph{Part (ii).}
Let $S_W$ be the writers in a wiring pattern $P \subseteq [w] \x [r]$, with $|S_W| \le d_k$.
Turn on only $S_W$, pick an injection $\iota: S_W \to [d_k]$, give each $i \in S_W$ the write key $k^m_{i} = e_{\iota(i)}$ ($k^m_{i} = 0$ for $i \notin S_W$), and give reader $\mathsf{R}_j$ the read query $q^m_{j} = \sum_{i \in S_W:\, (i, j) \in P} e_{\iota(i)}$.
Then $\langle q^m_{j}, k^m_{i} \rangle = 1$ if $(i, j) \in P$ and $0$ otherwise for all $i \in [w]$, so $c_{ij} \ne 0$ if and only if $(i, j) \in P$; since the virtual weights of the matrix stream are identities, the connected pairs are $P$, with one head per reader and $d_k$ independent of $w$.
The RMSNorm of the actual ports rescales each key and each query by a constant, and the sigmoid gates set their magnitudes (turning off writers outside $S_W$), so the pattern is realized up to per-reader scales.
In the full model, computing $\iota$ and $P$ from the vector stream is the role of the port generators on the control path.

\subsection{Wiring matrices of residual-stream mechanisms} \label{app:designs}
This section derives the entries of \cref{tab:design-space}.
As in \cref{framework}, we consider the wiring within a token from the earlier head writers ($o_i \in \R^d$) to the query, key and value readers ($u_j$) of one later layer.
Other reads of the history---the FFN inputs of HC/mHC and AttnRes, the residual input of MUDDFormer and the bypass $o^m$ of LoomFormer---feed no head input and are not readers.
We mark a port by $(x)$ if it is computed from the vector (residual) stream and by $(M)$ if it is computed from the memory itself.
Per-token scales such as pre-LN are absorbed into the wiring matrix as in \cref{lemma:fixed}, and the node-gating factors $g$, $g'$ of \cref{def:node-gated} are common to all mechanisms, so we omit them from the tables.
For LoomFormer we consider the matrix stream only and ignore MShift, which mixes neighbouring tokens (the $d/4$-dimensional RoPE slice from $x$ belongs to the node-gated vector stream of \cref{prop:rectangle}).
Ports are row vectors, as in \cref{affordable}.

\paragraph{A common form.}
Each mechanism keeps at a token a memory $M$ of $d_k$ slots of width $d_v$.
Writer $\mathsf{W}_i$ adds $k^m_{i} \otimes (o_i W^O_i)$ to $M$, reader $\mathsf{R}_j$ receives $u_j = (q^m_{j} M)\, W^s_j$, and between layers $M$ may be updated linearly, $M \leftarrow \Phi_l M$; here $k^m_{i}$ and $q^m_{j}$ are the write and read ports, and $W^O_i = W^s_j = I_d$ in the matrix stream of LoomFormer.
Composing these steps gives
\begin{equation} \label{eq:general-wiring}
    u_j = \sum_i \big(q^m_{j}\, \Phi_{l_i l_j}\, (k^m_{i})^\T\big)\, o_i W^O_i W^s_j,
\end{equation}
where $\Phi_{l_i l_j} = \Phi_{l_j - 1} \cdots \Phi_{l_i}$ is the product of the updates between the layer $l_i$ of $\mathsf{W}_i$ and the layer $l_j$ of $\mathsf{R}_j$: the wiring matrix has entries $c_{ij} = q^m_{j} \Phi_{l_i l_j} (k^m_{i})^\T$ and the virtual weights are $\bar C_{ij} = W^O_i W^s_j$.
This representation is not unique: a change of slot basis by an invertible $G$ maps $k^m_{i} \mapsto k^m_{i} G^\T$, $q^m_{j} \mapsto q^m_{j} G^{-1}$ and $\Phi_l \mapsto G \Phi_l G^{-1}$ without changing $c_{ij}$.
For mechanisms that keep the outputs of all earlier layers, we use the basis in which every stored output has its own slot, the write port is $e_{l_i}$ and $\Phi_l = I$; residual copies then become part of the read port.
Transitions $\Phi_{l_i l_j}$ can also be absorbed into the ports without adding slots: for a fixed reader layer $l_j$, setting $\tilde{k}^m_i = k^m_i \Phi^\T_{l_i l_j}$ writes $c_{ij} = \langle q^m_j, \tilde{k}^m_i \rangle$ even when $\Phi_l$ is singular (and when every $\Phi_l$ is invertible, setting $\Phi^\times_l = \Phi_{l-1} \cdots \Phi_1$ makes this split independent of $l_j$ via $\tilde{q}^m_j = q^m_j \Phi^\times_{l_j}$ and $\tilde{k}^m_i = k^m_i \Phi^{\times -\T}_{l_i}$).
\cref{tab:ports} lists the parameterization of each mechanism, justified below; \cref{tab:design-space} omits factors equal to $1$ or $I$.

\begin{table}[t]
\centering
\caption{Ports of residual-stream mechanisms in the common form of Eq.~(\ref{eq:general-wiring}).}
\label{tab:ports}
\begin{small}
\setlength{\tabcolsep}{4pt}
\begin{tabular}{lccc}
\toprule
 & Write port $k^m_i$ & Read port $q^m_j$ & Update $\Phi_l$ \\
\midrule
Transformer & $1$ & $1$ & $1$ \\
DenseFormer & $e_{l_i}$ & static & $I$ \\
HC / mHC & $\beta_{l_i}(M)$ & $\alpha_{l_j}(M)$ & $A_l(M)$ \\
AttnRes (full) & $\op{Norm}(v_{l_i})$ & static, softmax & $I$ \\
MUDDFormer & $e_{l_i}$ & per Q/K/V, from $x$ & $I$ \\
\midrule
LoomFormer & per head, from $x$ & per head and Q/K/V, from $x$ & $I$ \\
\bottomrule
\end{tabular}
\end{small}
\vskip -0.08in
\end{table}

\paragraph{Transformer.}
One slot of width $D$ with $k^m_{i} = q^m_{j} = \Phi = 1$ gives $c_{ij} = 1$ and the virtual weights $\bar C_{ij} = W^O_i W^s_j$; with the node gates, this is Eq.~(\ref{eq:node-gated}).

\paragraph{DenseFormer.}
The input of each block is a static weighted sum of the embedding and the outputs of all earlier blocks, and the block adds its own attention and FFN outputs to this input through its internal residual \cite{pagliardini2024denseformer}.
Unfolding these residual copies, the input of block $l'$ is a static combination of the outputs of all earlier blocks, so $c_{ij} \bar C_{ij} = \langle q^m_{l_j}, k^m_{l_i} \rangle\, W^O_i W^s_j$ with a static read port, so only node gating makes the wiring matrix input-dependent.
DenseFormer keeps one slot per block, about $l$ slots at layer $l$ (size $lD$).

\paragraph{HC and mHC.}
HC keeps $d_k$ parallel streams of width $D$ (we use $d_k = 4$, the main configuration of \citet{zhu2024hyper}).
Each sub-layer reads $\alpha M$ and updates the streams as $M \leftarrow A M + \beta^\T h$, with $\alpha$, $\beta$ and $A$ all computed from $M$ in the dynamic version \cite{zhu2024hyper}, and $A$ projected onto the Birkhoff polytope of doubly stochastic matrices in mHC \cite{xie2025mhc}.
Heads read and write through the attention sub-layer, with ports $\alpha_l$ and $\beta_l$, while the FFN and subsequent attention stream-mixing matrices between attention sub-layers $l_i$ and $l_j$ compose into $\Phi_{l_i l_j} = A_{l_i l_j}$.
In the common form, $c_{ij} = \alpha_{l_j} A_{l_i l_j} \beta_{l_i}^\T$---the unrolled weight of Eq.~(27) of \citet{zhu2024hyper}, which \citet{kimi2026attention} interpret as depth-wise linear attention with queries $\alpha$ and keys $\beta$.
Since all factors are computed from the streams $M$, absorbing $A_{l_i l_j}$ into the write port as above (without requiring $A_l$ to be invertible) gives the entry $\langle q^m_{l_j}(M), k^m_{l_i}(M) \rangle$ of \cref{tab:design-space}, with $d_k$-dimensional ports and state size $d_k D = 4D$.

\paragraph{AttnRes.}
Full AttnRes keeps the embedding and the outputs $v_m$ of all earlier sub-layers, without a residual connection, and gives sub-layer $m'$ the input $\sum_m \op{softmax}_m \langle w_{m'}, \op{Norm}(v_m) \rangle\, v_m$ with a learned query $w_{m'}$ \cite{kimi2026attention}.
In the slot basis of Eq.~(\ref{eq:general-wiring}), writers of layer $l_i$ write to their attention sub-layer's slot ($e_{l_i}$, while FFN and embedding slots hold no head writer) and heads of layer $l_j$ read linearly with the softmax weight vector over all stored slots; \cref{tab:design-space,tab:ports} instead display the underlying softmax query $q^m_{l_j} = w_{l_j}$ (static) and key $k^m_{l_i}(M) = \op{Norm}(v_{l_i})$ (from $M$), normalized by $\op{softmax}$ across all stored sub-layer slots.
Full AttnRes keeps one slot per sub-layer, about $2l$ slots at layer $l$ (size $2lD$); its block variant sums the outputs within each block into one slot.

\paragraph{MUDDFormer.}
MUDDFormer keeps the embedding and the outputs of all earlier blocks, and gives the next block four inputs---for its queries, keys, values and residual---each a weighted sum of these outputs with weights generated from the latest hidden state \cite{xiao2025muddformer}.
The residual input carries dynamic combinations of earlier outputs into the next block output; unfolding these recursions expresses the effective read port $q^{m,s}_{l_j}(x)$ of input $s \in \{Q, K, V\}$ as a function of the hidden states along the vector stream $x$, giving $c_{ij} \bar C_{ij} = \langle q^{m,s}_{l_j}(x), k^m_{l_i} \rangle\, W^O_i W^s_j$.
Like DenseFormer, MUDDFormer keeps about $l$ slots at layer $l$ (size $lD$).

\paragraph{LoomFormer.}
The matrix stream keeps $d_k$ slots of width $d$ that accumulate writes ($\Phi_l = I$) and passes content without fixed projections: absorbing the per-token RMS scales of $\op{Norm}(o_i)$ and $\op{MNorm}$ into $k^m_i$ and $q^{m,s}_j$ (and the learned diagonal gain of $\op{Norm}(o_i)$ into the writer output $o_i$) leaves $\bar C_{ij} = I_d$.
Indexing a reader by its head $j \in [n]$ and input $s \in \{Q, K, V\}$ as in \cref{tab:design-space} gives $c_{ij} = \langle q^{m,s}_{j}(x), k^m_{i}(x) \rangle$, with both ports generated per head and per token from the vector stream (\cref{prop:ports}(ii)).
With $d_k = 2n$, $M$ has size $d_k d = 2D$, plus $D$ for the vector stream.

\paragraph{Wiring ranks.}
Node gating multiplies $C$ entrywise by $g\, g'^\T$---equivalently $C \mapsto \op{diag}(g)\, C\, \op{diag}(g')$---which cannot increase its rank.
All readers of a layer share one read port in DenseFormer, HC/mHC and AttnRes, so $C$ has a single distinct column and rank at most $1$; MUDDFormer has one read port per Q/K/V input, hence at most three distinct columns and rank at most $3$ (reached once at least three earlier layers are present with linearly independent Q/K/V read ports).
In LoomFormer every reader has its own query, so the rank is that of $K Q^\T$ restricted to the $3n$ readers of the layer, at most $\min(w, d_k, 3n)$---hence $\min(d_k, 3n)$ once $w \ge \min(d_k, 3n)$, reached whenever the writer keys and reader queries span a common $\min(d_k, 3n)$-dimensional subspace.

\paragraph{Linear attention.}
The same form describes linear attention along the sequence instead of depth: the writers and readers are one head at different tokens, $M$ is the head's private state, the ports are its keys and queries, and the update $\Phi$ distinguishes the variants---$I$ for linear attention \cite{katharopoulos2020transformers}, a scalar decay in Mamba2 \cite{dao2024transformers}, a gated diagonal in gated linear attention \cite{yang2024gated}, $I - \beta k^\T k$ in DeltaNet \cite{yang2024parallelizing}, and $\gamma(I - \beta k^\T k)$ in Gated DeltaNet \cite{yang2025gated}.
It differs from LoomFormer in the axis and the sharing of the state (\cref{connections}).

\section{Extended Related Work} \label{related_work}
\paragraph{Cross-layer connections and residual-stream architectures.}
Beyond standard residual connections \cite{he2016deep}, early work strengthened depth-wise signal propagation through input-dependent gating in Highway Networks \cite{srivastava2015training} and all-to-all layer aggregation with static or learnable weights in DenseNet \cite{huang2017densely}, DLCL \cite{wang2019learning} and DenseFormer \cite{pagliardini2024denseformer}.
Subsequent architectures introduced cross-layer attention over earlier layer representations \cite{elnokrashy2022depth,fang2023cross,wang2024strengthening,kimi2026attention}, multi-stream residual states with dynamic or manifold-constrained mixing in HC and mHC \cite{zhu2024hyper,xie2025mhc}, and decoupled Q/K/V/residual layer aggregation in MUDDFormer \cite{xiao2025muddformer}.
As analyzed in \cref{connections} and \cref{app:designs}, all of these mechanisms operate at the granularity of whole layers (or layer $\x$ Q/K/V), summing the outputs of all heads within a layer before routing and restricting the within-token wiring rank to $1$ (or $3$).

The Residual Matrix Transformer (RMT) \cite{mak2025residual} replaces the vector residual stream entirely with a pure matrix residual stream, which heads and FFNs read and write through static key vectors so the stream can be widened cheaply.
LoomFormer instead keeps a dual-stream architecture that separates control on the vector stream from content on the matrix stream: the compact vector stream generates per-token ports cheaply, whereas RMT's static ports fall under \cref{prop:ports}(i) and remain node-gated however wide the matrix is made ($4D$ in RMT's main configuration vs.\ $2D + D$ in LoomFormer).

\paragraph{Cross-layer token attention and KV sharing.}
OmniNet \cite{tay2021omninet} lets every token attend directly to all tokens across all layers, incurring attention cost quadratic in $L T$; loom attention instead factors cross-layer routing within each token through a fixed-size matrix memory $M$ in $\mathcal{O}(L T D d_k)$ time and composes it with standard within-layer horizontal attention.
Cross-Layer Attention (CLA) \cite{brandon2024reducing} shares key-value caches across adjacent layers to reduce inference memory at a modest quality trade-off, whereas loom attention dynamically composes earlier head outputs into distinct per-head queries, keys and values at every layer (\cref{analyzing}).

\paragraph{Mechanistic interpretability and intra-layer advances.}
Circuit-level interpretability studies \cite{Elhage2021mathematical,olsson2022context,conmy2023towards,hanna2024have,merullo2024talking,ni2024benchmarking} show that Transformers implement tasks by composing specific upstream and downstream heads across layers; loom attention turns this implicit, static composition into an explicit, per-token operation over $M$.
Orthogonal intra-layer advances---including improved horizontal attention \cite{liu2024deepseek,xiao2024improving,ye2024differential,leviathan2024selective}, sequence-axis linear attention and state-space models \cite{gu2023mamba,dao2024transformers,peng2024eagle,yang2024parallelizing,yang2025gated}, and sparse Mixture-of-Experts FFNs \cite{fedus2022switch,dai2024deepseekmoe}---modify horizontal mixing or FFN capacity within a layer and can be combined directly with loom attention along depth.

\section{PyTorch-Style Pseudocode for LoomFormer}
\label{pseudo code}
\begin{Verbatim}[commandchars=\\\{\},frame=lines,framesep=2mm,baselinestretch=1.15,fontsize=\footnotesize,breaklines,numbersep=6pt,numbers=left]
\PYG{c+c1}{\PYGZsh{} b = batch size, t = seq len, n = num\PYGZus{}heads, d = head\PYGZus{}dim}
\PYG{c+c1}{\PYGZsh{} k = num\PYGZus{}slots d\PYGZus{}k, r = RoPE dim d // 4. M is of shape btkd}

\PYG{k}{def} \PYG{n+nf}{MRead}\PYG{p}{(}\PYG{n}{M}\PYG{p}{,} \PYG{n}{x}\PYG{p}{,} \PYG{n}{xr}\PYG{p}{):}
    \PYG{n}{qm} \PYG{o}{=} \PYG{n}{rearrange}\PYG{p}{(}\PYG{n}{xr} \PYG{o}{@} \PYG{n}{W\PYGZus{}r2}\PYG{p}{,} \PYG{l+s+s1}{\PYGZsq{}bt(nk)\PYGZhy{}\PYGZgt{}btnk\PYGZsq{}}\PYG{p}{)}
    \PYG{n}{qm} \PYG{o}{=} \PYG{n}{rmsnorm}\PYG{p}{(}\PYG{n}{qm}\PYG{p}{)} \PYG{o}{*} \PYG{n}{sigmoid}\PYG{p}{(}\PYG{n}{rearrange}\PYG{p}{(}\PYG{n}{x} \PYG{o}{@} \PYG{n}{W\PYGZus{}rg}\PYG{p}{,} \PYG{l+s+s1}{\PYGZsq{}btn\PYGZhy{}\PYGZgt{}btn1\PYGZsq{}}\PYG{p}{))}
    \PYG{n}{qm} \PYG{o}{+=} \PYG{n}{q\PYGZus{}bar} \PYG{o}{*} \PYG{n}{sigmoid}\PYG{p}{(}\PYG{n}{rearrange}\PYG{p}{(}\PYG{n}{x} \PYG{o}{@} \PYG{n}{W\PYGZus{}rg\PYGZus{}bar}\PYG{p}{,} \PYG{l+s+s1}{\PYGZsq{}btn\PYGZhy{}\PYGZgt{}btn1\PYGZsq{}}\PYG{p}{))} \PYG{c+c1}{\PYGZsh{} nk,btn1\PYGZhy{}\PYGZgt{}btnk}
    \PYG{n}{u} \PYG{o}{=} \PYG{n}{einsum}\PYG{p}{(}\PYG{l+s+s1}{\PYGZsq{}btnk,btkd\PYGZhy{}\PYGZgt{}btnd\PYGZsq{}}\PYG{p}{,} \PYG{n}{qm}\PYG{p}{,} \PYG{n}{M}\PYG{p}{)}
    \PYG{k}{return} \PYG{n}{u}
    
\PYG{k}{def} \PYG{n+nf}{MWrite}\PYG{p}{(}\PYG{n}{o}\PYG{p}{,} \PYG{n}{x}\PYG{p}{,} \PYG{n}{xw}\PYG{p}{):}
    \PYG{n}{km} \PYG{o}{=} \PYG{n}{rearrange}\PYG{p}{(}\PYG{n}{xw} \PYG{o}{@} \PYG{n}{W\PYGZus{}w2}\PYG{p}{,} \PYG{l+s+s1}{\PYGZsq{}bt(nk)\PYGZhy{}\PYGZgt{}btnk\PYGZsq{}}\PYG{p}{)}
    \PYG{n}{km} \PYG{o}{=} \PYG{n}{rmsnorm}\PYG{p}{(}\PYG{n}{km}\PYG{p}{)} \PYG{o}{*} \PYG{n}{sigmoid}\PYG{p}{(}\PYG{n}{rearrange}\PYG{p}{(}\PYG{n}{x} \PYG{o}{@} \PYG{n}{W\PYGZus{}wg}\PYG{p}{,} \PYG{l+s+s1}{\PYGZsq{}btn\PYGZhy{}\PYGZgt{}btn1\PYGZsq{}}\PYG{p}{))}
    \PYG{n}{dM} \PYG{o}{=} \PYG{n}{einsum}\PYG{p}{(}\PYG{l+s+s1}{\PYGZsq{}btnk,btnd\PYGZhy{}\PYGZgt{}btkd\PYGZsq{}}\PYG{p}{,} \PYG{n}{km}\PYG{p}{,} \PYG{n}{o}\PYG{p}{)}
    \PYG{k}{return} \PYG{n}{dM}
    
\PYG{k}{def} \PYG{n+nf}{MShift}\PYG{p}{(}\PYG{n}{M}\PYG{p}{,} \PYG{n}{x}\PYG{p}{):}
    \PYG{n}{M\PYGZus{}prev} \PYG{o}{=} \PYG{n}{shift\PYGZus{}right}\PYG{p}{(}\PYG{n}{M}\PYG{p}{,} \PYG{l+m+mi}{1}\PYG{p}{,} \PYG{n}{dim}\PYG{o}{=}\PYG{l+m+mi}{1}\PYG{p}{)}  \PYG{c+c1}{\PYGZsh{} shift t dim of btkd}
    \PYG{n}{z} \PYG{o}{=} \PYG{n}{sigmoid}\PYG{p}{(}\PYG{n}{rearrange}\PYG{p}{(}\PYG{n}{x} \PYG{o}{@} \PYG{n}{W\PYGZus{}sg}\PYG{p}{,} \PYG{l+s+s1}{\PYGZsq{}btk\PYGZhy{}\PYGZgt{}btk1\PYGZsq{}}\PYG{p}{))}
    \PYG{k}{return} \PYG{n}{z} \PYG{o}{*} \PYG{n}{M\PYGZus{}prev} \PYG{o}{+} \PYG{p}{(}\PYG{l+m+mi}{1} \PYG{o}{\PYGZhy{}} \PYG{n}{z}\PYG{p}{)} \PYG{o}{*} \PYG{n}{M}
    
\PYG{k}{def} \PYG{n+nf}{loom\PYGZus{}mha}\PYG{p}{(}\PYG{n}{x}\PYG{p}{,} \PYG{n}{M}\PYG{p}{,} \PYG{n}{write\PYGZus{}M}\PYG{o}{=}\PYG{k+kc}{True}\PYG{p}{):}
    \PYG{n}{M} \PYG{o}{=} \PYG{n}{RMSNorm}\PYG{p}{(}\PYG{n}{MShift}\PYG{p}{(}\PYG{n}{M}\PYG{p}{,} \PYG{n}{x}\PYG{p}{),} \PYG{n}{dim}\PYG{o}{=}\PYG{p}{(}\PYG{o}{\PYGZhy{}}\PYG{l+m+mi}{2}\PYG{p}{,} \PYG{o}{\PYGZhy{}}\PYG{l+m+mi}{1}\PYG{p}{))}
    \PYG{n}{xr}\PYG{p}{,} \PYG{n}{xw} \PYG{o}{=} \PYG{n}{gelu}\PYG{p}{(}\PYG{n}{x} \PYG{o}{@} \PYG{n}{W\PYGZus{}r1}\PYG{p}{),} \PYG{n}{gelu}\PYG{p}{(}\PYG{n}{x} \PYG{o}{@} \PYG{n}{W\PYGZus{}w1}\PYG{p}{)}
    \PYG{n}{q}\PYG{p}{,} \PYG{n}{k}\PYG{p}{,} \PYG{n}{v}\PYG{p}{,} \PYG{n}{o} \PYG{o}{=} \PYG{p}{[}\PYG{n}{MRead}\PYG{p}{[}\PYG{n}{s}\PYG{p}{](}\PYG{n}{M}\PYG{p}{,} \PYG{n}{x}\PYG{p}{,} \PYG{n}{xr}\PYG{p}{)} \PYG{k}{for} \PYG{n}{s} \PYG{o+ow}{in} \PYG{p}{[}\PYG{l+s+s1}{\PYGZsq{}q\PYGZsq{}}\PYG{p}{,} \PYG{l+s+s1}{\PYGZsq{}k\PYGZsq{}}\PYG{p}{,} \PYG{l+s+s1}{\PYGZsq{}v\PYGZsq{}}\PYG{p}{,} \PYG{l+s+s1}{\PYGZsq{}o\PYGZsq{}}\PYG{p}{]]}
    \PYG{n}{qr} \PYG{o}{=} \PYG{n}{rope}\PYG{p}{(}\PYG{n}{rearrange}\PYG{p}{(}\PYG{n}{x} \PYG{o}{@} \PYG{n}{W\PYGZus{}q}\PYG{p}{,} \PYG{l+s+s1}{\PYGZsq{}bt(nr)\PYGZhy{}\PYGZgt{}btnr\PYGZsq{}}\PYG{p}{))}
    \PYG{n}{kr} \PYG{o}{=} \PYG{n}{rope}\PYG{p}{(}\PYG{n}{rearrange}\PYG{p}{(}\PYG{n}{x} \PYG{o}{@} \PYG{n}{W\PYGZus{}k}\PYG{p}{,} \PYG{l+s+s1}{\PYGZsq{}bt(nr)\PYGZhy{}\PYGZgt{}btnr\PYGZsq{}}\PYG{p}{))}
    \PYG{n}{q}\PYG{p}{,} \PYG{n}{k} \PYG{o}{=} \PYG{n}{concat}\PYG{p}{([}\PYG{n}{qr}\PYG{p}{,} \PYG{n}{q}\PYG{p}{],} \PYG{n}{dim}\PYG{o}{=\PYGZhy{}}\PYG{l+m+mi}{1}\PYG{p}{),} \PYG{n}{concat}\PYG{p}{([}\PYG{n}{kr}\PYG{p}{,} \PYG{n}{k}\PYG{p}{],} \PYG{n}{dim}\PYG{o}{=\PYGZhy{}}\PYG{l+m+mi}{1}\PYG{p}{)}

    \PYG{n}{allow\PYGZus{}mask} \PYG{o}{=} \PYG{n}{tril}\PYG{p}{(}\PYG{n}{ones}\PYG{p}{(}\PYG{n}{T}\PYG{p}{,} \PYG{n}{T}\PYG{p}{),} \PYG{n}{diagonal}\PYG{o}{=\PYGZhy{}}\PYG{l+m+mi}{1}\PYG{p}{)} \PYG{c+c1}{\PYGZsh{} strictly causal mask}
    \PYG{n}{o} \PYG{o}{=} \PYG{n}{o} \PYG{o}{+} \PYG{n}{MHA}\PYG{p}{(}\PYG{n}{q}\PYG{p}{,} \PYG{n}{k}\PYG{p}{,} \PYG{n}{v}\PYG{p}{,} \PYG{n}{allow\PYGZus{}mask}\PYG{o}{=}\PYG{n}{allow\PYGZus{}mask}\PYG{p}{)}
    \PYG{n}{dx} \PYG{o}{=} \PYG{n}{rearrange}\PYG{p}{(}\PYG{n}{o}\PYG{p}{,} \PYG{l+s+s1}{\PYGZsq{}btnd\PYGZhy{}\PYGZgt{}bt(nd)\PYGZsq{}}\PYG{p}{)} \PYG{o}{@} \PYG{n}{W\PYGZus{}o}
    \PYG{n}{dM} \PYG{o}{=} \PYG{n}{MWrite}\PYG{p}{(}\PYG{n}{RMSNorm}\PYG{p}{(}\PYG{n}{o}\PYG{p}{,} \PYG{n}{dim}\PYG{o}{=\PYGZhy{}}\PYG{l+m+mi}{1}\PYG{p}{),} \PYG{n}{x}\PYG{p}{,} \PYG{n}{xw}\PYG{p}{)} \PYG{k}{if} \PYG{n}{write\PYGZus{}M} \PYG{k}{else} \PYG{l+m+mf}{0.}
    \PYG{k}{return} \PYG{n}{dx}\PYG{p}{,} \PYG{n}{dM}
    
\PYG{k}{def} \PYG{n+nf}{ffn}\PYG{p}{(}\PYG{n}{x}\PYG{p}{,} \PYG{n}{write\PYGZus{}M}\PYG{o}{=}\PYG{k+kc}{True}\PYG{p}{):}
    \PYG{n}{xw} \PYG{o}{=} \PYG{n}{gelu}\PYG{p}{(}\PYG{n}{x} \PYG{o}{@} \PYG{n}{W\PYGZus{}w1}\PYG{p}{)}
    \PYG{n}{dx} \PYG{o}{=} \PYG{n}{\PYGZus{}ffn\PYGZus{}x}\PYG{p}{(}\PYG{n}{x}\PYG{p}{)}
    \PYG{n}{dM} \PYG{o}{=} \PYG{n}{MWrite}\PYG{p}{(}\PYG{n}{rearrange}\PYG{p}{(}\PYG{n}{\PYGZus{}ffn\PYGZus{}m}\PYG{p}{(}\PYG{n}{x}\PYG{p}{),} \PYG{l+s+s1}{\PYGZsq{}bt(nd)\PYGZhy{}\PYGZgt{}btnd\PYGZsq{}}\PYG{p}{),} \PYG{n}{x}\PYG{p}{,} \PYG{n}{xw}\PYG{p}{)} \PYG{k}{if} \PYG{n}{write\PYGZus{}M} \PYG{k}{else} \PYG{l+m+mf}{0.}
    \PYG{k}{return} \PYG{n}{dx}\PYG{p}{,} \PYG{n}{dM}
    
\PYG{k}{def} \PYG{n+nf}{loomformer}\PYG{p}{(}\PYG{n}{input\PYGZus{}ids}\PYG{p}{):}
    \PYG{n}{x} \PYG{o}{=} \PYG{n}{embedding}\PYG{p}{(}\PYG{n}{input\PYGZus{}ids}\PYG{p}{)}
    \PYG{n}{o} \PYG{o}{=} \PYG{n}{rearrange}\PYG{p}{(}\PYG{n}{x} \PYG{o}{@} \PYG{n}{W\PYGZus{}e}\PYG{p}{,} \PYG{l+s+s1}{\PYGZsq{}bt(nd)\PYGZhy{}\PYGZgt{}btnd\PYGZsq{}}\PYG{p}{)}
    \PYG{n}{M} \PYG{o}{=} \PYG{n}{MWrite}\PYG{p}{(}\PYG{n}{o}\PYG{p}{,} \PYG{n}{x}\PYG{p}{,} \PYG{n}{gelu}\PYG{p}{(}\PYG{n}{x} \PYG{o}{@} \PYG{n}{W\PYGZus{}w1}\PYG{p}{))}

    \PYG{k}{for} \PYG{n}{l} \PYG{o+ow}{in} \PYG{n+nb}{range}\PYG{p}{(}\PYG{n}{L}\PYG{p}{):}
        \PYG{n}{write\PYGZus{}M} \PYG{o}{=} \PYG{n}{l} \PYG{o}{\PYGZlt{}} \PYG{n}{L} \PYG{o}{\PYGZhy{}} \PYG{l+m+mi}{1}
        \PYG{c+c1}{\PYGZsh{} pairs are added element\PYGZhy{}wise: (x, M) + (dx, dM) = (x + dx, M + dM)}
        \PYG{p}{(}\PYG{n}{x}\PYG{p}{,} \PYG{n}{M}\PYG{p}{)} \PYG{o}{=} \PYG{p}{(}\PYG{n}{x}\PYG{p}{,} \PYG{n}{M}\PYG{p}{)} \PYG{o}{+} \PYG{n}{loom\PYGZus{}mha}\PYG{p}{[}\PYG{n}{l}\PYG{p}{](}\PYG{n}{ln\PYGZus{}1}\PYG{p}{[}\PYG{n}{l}\PYG{p}{](}\PYG{n}{x}\PYG{p}{),} \PYG{n}{M}\PYG{p}{,} \PYG{n}{write\PYGZus{}M}\PYG{o}{=}\PYG{n}{write\PYGZus{}M}\PYG{p}{)}
        \PYG{p}{(}\PYG{n}{x}\PYG{p}{,} \PYG{n}{M}\PYG{p}{)} \PYG{o}{=} \PYG{p}{(}\PYG{n}{x}\PYG{p}{,} \PYG{n}{M}\PYG{p}{)} \PYG{o}{+} \PYG{n}{ffn}\PYG{p}{[}\PYG{n}{l}\PYG{p}{](}\PYG{n}{ln\PYGZus{}2}\PYG{p}{[}\PYG{n}{l}\PYG{p}{](}\PYG{n}{x}\PYG{p}{),} \PYG{n}{write\PYGZus{}M}\PYG{o}{=}\PYG{n}{write\PYGZus{}M}\PYG{p}{)}
    \PYG{k}{return} \PYG{n}{unembedding}\PYG{p}{(}\PYG{n}{ln\PYGZus{}f}\PYG{p}{(}\PYG{n}{x}\PYG{p}{))}
\end{Verbatim}

\section{Experimental Configurations, Baselines and Benchmarks} \label{hyperparameters}
\subsection{Model Configurations and Hyperparameters} \label{hparams for lm}
\cref{tab:scaling-configs} lists the architectural dimensions, peak learning rates and token budgets across all four LoomFormer scales (\cref{scalinglaws,downstream}).
In blocks $0, \dots, L - 2$, $\op{FFN}_x$ and the $\op{FFN}_M$ write branch ($\op{FFN}_M$ plus its MWrite module) follow a $2{:}1$ parameter ratio (rounded to multiples of $128$ at 2.8B, giving $2.2{:}1$), while block $L - 1$ (last) omits unused writes to $M_L$ and widens $\op{FFN}_x$ to match the baseline Transformer's total parameter count.
\cref{tab:scaling-hparams} lists the training hyperparameters for the scaling-law experiments (\cref{scalinglaws}; weight decay is excluded on normalization scales and biases, including gate biases), while the 2.8B models (\cref{downstream}) follow Pythia-2.8B \citep[Appendix~E]{biderman2023pythia}.
To improve training stability at the XL scale, LoomFormer-XL replaces SwiGLU in $\op{FFN}_x$ with Kimi K3's bounded SiTU-GLU activation \cite{kimi2026k3} with $\beta_1 = 4$, $\beta_2 = 7$.

\paragraph{Port and gate initialization.}
Each gate projection in Eqs.~(\ref{eq:MRead})--(\ref{eq:loom-read}) includes a learnable bias (omitted in \cref{architecture} for brevity) that sets its initial opening.
Specifically, the gate projection weights are initialized to zero, with biases setting initial gate values to $0.05$ on the dynamic read branch, $0.9$ on the static read branch, $0.1$ on the write branch, and $0.5$ on MShift ($z_{[t]}$).
For the static queries $\bar{q} \in \R^{d_k}$, value reads $v$ are initialized from $\mathcal{N}(0, 1/d_k)$, whereas $\tilde{q}, \tilde{k}$ and $o^m$ start at $\bar{q} = 0$ so memory-read Q/K coordinates and the output bypass grow in gradually.

\begin{table}[htb]
\centering
\caption{Model configurations across scales. $d_k = 2n$; shared port-feature width $D_r = n d_k / 2 = n^2$, capped at $D/4 = 640$ at 2.8B.}
\label{tab:scaling-configs}
\begin{small}
\setlength{\tabcolsep}{4.5pt}
\begin{tabular}{lcccccccccccc}
\toprule
Model & Params & $L$ & $D$ & $D/L$ & $n$ & $d$ & $D_r$ & Tfm FFN & $\op{FFN}_x$ (last) & $\op{FFN}_M$ & LR & Tokens \\
\midrule
Medium & 432M & 18 & 1200 & 66.7 & 16 & 75 & 256 & 3200  & 2510 (3904) & 1128 & 3.0e-4 & 7B \\
Large  & 778M & 22 & 1536 & 69.8 & 16 & 96 & 256 & 4096  & 3317 (5083) & 1539 & 2.5e-4 & 14B \\
XL     & 1.4B & 28 & 1920 & 68.6 & 20 & 96 & 400 & 5120  & 4048 (6277) & 1827 & 2.0e-4 & 26B \\
2.8B   & 2.8B & 32 & 2560 & 80   & 32 & 80 & 640 & 10240 & 7808 (9984) & 2944 & 1.6e-4 & 300B \\
\bottomrule
\end{tabular}
\end{small}
\end{table}

\begin{table}[htb]
\centering
\caption{Training hyperparameters for the scaling-law experiments (\cref{scalinglaws}).}
\label{tab:scaling-hparams}
\begin{small}
\setlength{\tabcolsep}{7pt}
\begin{tabular}{ll @{\hspace{24pt}} ll}
\toprule
Hyperparameter & Value & Hyperparameter & Value \\
\midrule
Optimizer            & AdamW                     & Sequence length   & $4096$ \\
$(\beta_1, \beta_2)$ & $(0.9, 0.95)$             & Batch size (seqs) & $128$ \\
$\epsilon$           & $10^{-8}$                 & Vocab size        & $50{,}432$ \\
Weight decay         & $0.1$                     & Tied embeddings   & False \\
Gradient clipping    & $1.0$                     & Linear bias       & False \\
LR schedule          & Cosine                    & Normalization     & RMSNorm ($\epsilon = 10^{-6}$) \\
Min LR ratio         & $0.1$                     & FFN activation    & SwiGLU \\
Warmup fraction      & $0.01$                    & Positional emb.   & RoPE ($\theta = 10{,}000$) \\
Dropout              & $0.0$                     & Precision         & BF16 \\
\bottomrule
\end{tabular}
\end{small}
\end{table}

\subsection{Scaling-Law Baselines} \label{baselines for lm}
All baselines in \cref{scalinglaws} are trained in JAX/MaxText on the identical data order, tokenizer and schedule as LoomFormer:
\begin{itemize}
    \item \textbf{Transformer++}: Llama-style pre-norm Transformer \cite{touvron2023llama} with RoPE \cite{su2024roformer}, SwiGLU FFN \cite{shazeer2020glu}, RMSNorm and no linear biases.
    \item \textbf{DHC}: Dynamic Hyper-Connections with expansion rate $d_k = 4$ and $\tanh$-bounded dynamic routing following \citet{zhu2024hyper}.
    \item \textbf{mHC}: Manifold-constrained Hyper-Connections ($d_k = 4$ streams) projecting the stream-mixing matrix onto the Birkhoff polytope via Sinkhorn normalization \cite{xie2025mhc}.
    \item \textbf{AttnRes (full)}: Full Attention Residuals aggregating all preceding sub-layer outputs per sub-layer via learned depth-wise softmax queries \cite{kimi2026attention}.
    \item \textbf{MUDDFormer}: Multiway Dynamic Dense connections generating per-token depth-aggregation weights separately for the query, key, value and residual inputs of each block \cite{xiao2025muddformer}.
\end{itemize}

\subsection{Downstream Evaluation Benchmarks} \label{app:downstream-datasets}
In addition to the benchmarks used by Pythia \cite{biderman2023pythia}---LAMBADA \cite{paperno2016lambada}, PIQA \cite{bisk2020piqa}, WinoGrande \cite{sakaguchi2021winogrande}, ARC-Easy/Challenge \cite{clark2018think} and SciQ \cite{welbl2017crowdsourcing}---we evaluate BoolQ \cite{clark2019boolq} and HellaSwag \cite{zellers2019hellaswag} for commonsense reasoning, and RACE-Middle/High \cite{lai2017race} for reading comprehension (\cref{tab:downstream}), all under LM Evaluation Harness \cite{eval-harness} in zero-shot and five-shot settings.

\section{Details of Complexity Analysis}
\label{append:complexity analysis}
We derive the forward-FLOP overhead in \cref{tab:complexty analysis} under a matched parameter budget, following the counting convention of DCFormer \cite{xiao2024improving} and MUDDFormer \cite{xiao2025muddformer}.
Let $L$ be the number of layers, $T$ the sequence length, $D$ the model width, $n$ the number of heads, $d=D/n$ the head dimension, and $M\in\R^{d_k\x d}$ the matrix memory.
We report FLOPs per sequence at two FLOPs per multiply-add, omitting the shared token embedding and unembedding as well as normalization, nonlinearities, biases and other elementwise operations (such as gate activations and the elementwise mixing in MShift).
Causal attention has approximately $T^2/2$ query--key pairs; using a strictly rather than inclusively causal mask changes only the lower-order diagonal terms.

\paragraph{Parameter budget and tokenwise projections.}
The reference Transformer has $4D^2$ attention and $8D^2$ FFN weights per layer, for a total of $P_0=12LD^2$ (applying to both a GELU FFN of hidden width $4D$ and a SwiGLU FFN of hidden width $8D/3$).
LoomFormer replaces the full Q/K/V projections with smaller RoPE projections and matrix reads, redistributing the saved parameter budget among the port generators, $\op{FFN}_x$ and $\op{FFN}_M$, as well as the initial value projection $W^E$ and embedding MWrite (Eq.~\ref{eq:loom-embed}), so that $P_{\mathrm{Loom}}=P_0$.

Every tokenwise dense matrix $W$ costs $2T|W|$ FLOPs, where $|W|$ is its number of entries.
Consequently, to leading order, the combined projection and FFN cost is $2TP_0$ in both models.
This includes both stages of the low-rank port generators, all projected gates (including $W^S_g$ in MShift), $W^E$, $\op{FFN}_x$ and $\op{FFN}_M$; a shared projection is counted once.
Adding the port-generation or $\op{FFN}_M$ FLOPs without subtracting the $\op{FFN}_x$ FLOPs that paid for their weights would double-count them.
Static ports, learned normalization scales and biases give only lower-order corrections to this dense-weight relation, and for the same reason the bottleneck width $D_r$ affects parameter allocation without entering the leading net FLOP overhead.

\paragraph{Reference Transformer FLOPs.}
In addition to the tokenwise projections and FFNs, each layer computes $QK^\T$ and attention-weighted values.
Each costs $2nd\cdot T^2/2=DT^2$ FLOPs under the causal approximation.
The reference forward cost is therefore
\begin{equation} \label{eq:complexity-baseline}
F_0 = 2TP_0+2LDT^2 = 2LDT(12D+T).
\end{equation}

\paragraph{Reading and writing the matrix memory.}
A single MRead contracts a $d_k$-dimensional port with $M\in\R^{d_k\x d}$, at a cost of $2d_kd$ FLOPs per token.
Its dynamic and static ports are summed before this contraction (Eq.~\ref{eq:MRead}), so they do not require two separate reads.
With four reads per head and $n$ heads, the $L$ layers cost
\begin{equation}
F_{\mathrm{read}}=4\cdot 2Tndd_k\cdot L=8LTDd_k.
\end{equation}
For each writing sub-layer, MWrite sums $n$ outer products of shape $d_k\x d$, costing $2Tndd_k=2TDd_k$ FLOPs.
There are two such sub-layers per block, attention and $\op{FFN}_M$, except in the last block, which does not write to $M$.
The embedding adds one initial write of the same size (Eq.~\ref{eq:loom-embed}).
Hence
\begin{equation} \label{eq:complexity-memory}
\begin{split}
F_{\mathrm{write}} &= 4(L-1)TDd_k+2TDd_k,\\
F_{\mathrm{M}} &= F_{\mathrm{read}}+F_{\mathrm{write}} = (12L-2)TDd_k\simeq 12LTDd_k.
\end{split}
\end{equation}
The last approximation neglects the small endpoint factor $1-1/(6L)$ multiplying the memory-overhead term.
Since $M$ has a fixed size, the read/write cost per layer does not increase with depth.

\paragraph{Wider query and key.}
In the reference Transformer, Q, K and V each have $d$ coordinates per head.
LoomFormer concatenates $d/4$ RoPE coordinates to the $d$ coordinates of Q/K read from $M$ (Eq.~\ref{eq:loom-read}), while V retains width $d$.
Thus only the $QK^\T$ contraction is wider; the attention-weighted value contraction is unchanged.
The extra cost is
\begin{equation} \label{eq:complexity-qk}
\Delta F_{\mathrm{QK}}
=2Ln\frac{d}{4}\cdot\frac{T^2}{2}=\frac{LDT^2}{4}.
\end{equation}

\paragraph{Total overhead and scaling.}
Combining Eqs.~(\ref{eq:complexity-baseline})--(\ref{eq:complexity-qk}) gives
\begin{equation} \label{eq:complexity-general}
R_{\Delta\mathrm{FLOPs}}
\simeq\frac{12LTDd_k+LDT^2/4}{2LDT(12D+T)} =\frac{d_k}{2D(1+\rho/12)}
 +\frac{1}{8(1+12/\rho)},
\end{equation}
where $\rho=T/D$.
With $D=nd$ and $d_k=2n$, we have $d_k/(2D)=1/d$, giving the expression in \cref{overhead}:
\begin{equation} \label{eq:complexity-simplified}
\boxed{
R_{\Delta\mathrm{FLOPs}}\simeq
\underbrace{\frac{1}{d(1+\rho/12)}}_{\text{MRead/MWrite}}
+\underbrace{\frac{1}{8(1+12/\rho)}}_{\text{wider Q/K}}.
}
\end{equation}
At fixed $\rho$, the matrix read/write term scales as $1/d$ while the wider-Q/K term is independent of $d$; as $\rho$ grows, the first term falls from $1/d$ toward $0$ and the second rises toward $1/8$, with both independent of $L$ at leading order.
In \cref{tab:complexty analysis} ($d = 75, 96, 96, 80$ and $\rho = T/D = 3.41, 2.67, 2.13, 0.80$ for Medium, Large, XL and 2.8B, using $T = 4096$ at Medium/Large/XL and $T = 2048$ at 2.8B; \cref{tab:scaling-configs}), Large, XL and 2.8B have integer $d/4$ and follow Eq.~(\ref{eq:complexity-simplified}) directly ($3.13\%$, $2.77\%$ and $1.95\%$), whereas Medium rounds its RoPE width $d/4 = 75/4$ to $\lfloor d/4 \rfloor = 18$ in Eq.~(\ref{eq:complexity-qk}), giving $3.70\%$ instead of Eq.~(\ref{eq:complexity-simplified})'s unrounded $3.81\%$ (retaining $12L - 2$ in Eq.~\ref{eq:complexity-memory} changes every row by $< 0.01$ percentage points).
For a backbone with a different FFN parameter ratio, one can retain the general denominator $2TP_0+2LDT^2$ instead of substituting $P_0=12LD^2$.
Wall-clock training and inference speeds are detailed in \cref{mem_usage}.

\section{Training and Inference Throughput Details} \label{mem_usage}

\paragraph{Training throughput.}
\cref{tab:train-overhead-full} reports paired training benchmarks in JAX/MaxText on TPU v5p pods at each model's training configuration (\cref{scalinglaws,downstream}), with both Transformer and LoomFormer using Splash attention ($1024$-token blocks and fused backward).
LoomFormer executes custom Pallas kernels (forward and backward) that fuse MShift, MNorm, port normalization and gating, and the four MReads of all $n$ heads into one per-token $(4n \x d_k) \x (d_k \x d)$ MXU matrix multiplication in VMEM, and likewise fuse MWrite's $n$-head outer-product accumulation, avoiding intermediate HBM traffic on $M$.
We compare median XPlane device step time under full activation rematerialization on both sides (\cref{tab:complexty analysis}).
At 2.8B, LoomFormer uses $512$-token cross-entropy chunking to fit \texttt{v5p-128} HBM whereas Pythia-2.8B does not, making the 2.8B ratio slightly conservative.

\begin{table}[htb]
\centering
\caption{TPU v5p training step time and relative throughput of LoomFormer (Loom) vs.\ Transformer (Tfm).}
\label{tab:train-overhead-full}
\begin{small}
\setlength{\tabcolsep}{6pt}
\begin{tabular}{llcc ccc}
\toprule
Model & Pod & $T$ & Batch (seqs) & Tfm (ms) & Loom (ms) & Relative speed \\
\midrule
Medium & v5p-16  & 4096 & 128  & $1224$ & $1352$ & $90.5\%$ \\
Large  & v5p-32  & 4096 & 128  & $962$  & $1078$ & $89.2\%$ \\
XL     & v5p-32  & 4096 & 128  & $1590$ & $1723$ & $92.3\%$ \\
2.8B   & v5p-128 & 2048 & 1024 & $2988$ & $3233$ & $92.4\%$ \\
\bottomrule
\end{tabular}
\end{small}
\end{table}

\paragraph{Inference throughput on NVIDIA A800.}
\cref{tab:infer-overhead-full} reports BF16 inference throughput on a single NVIDIA A800-80GB PCIe GPU (\texttt{torch} 2.14 / \texttt{triton} 3.8 / cuDNN 9.24) using trained checkpoints at all four scales, taking the median over five measurements.
LoomFormer executes one fused Triton kernel per layer covering the preceding write, the MShift read view, the four MReads, RoPE and cache updates, together with grouped decode GEMVs for the small port projections; prefill uses each model's fastest SDPA backend (FlashAttention for Transformer; cuDNN SDPA for LoomFormer, whose Q/K head width $1.25d$ differs from V width $d$).
Prefill measures one uncached forward pass of $B \x T$ tokens, and decode measures single-token generation steps after a $(T - 64)$-token prompt under \texttt{torch.compile(mode="max-autotune")} with CUDA graphs.

\begin{table}[htb]
\centering
\caption{BF16 inference throughput on one NVIDIA A800-80GB GPU, reported as relative throughput and absolute tokens/s (\texttt{Loom / Tfm}); $T = 4096$ for Medium, Large and XL, and $T = 2048$ for 2.8B.}
\label{tab:infer-overhead-full}
\begin{small}
\setlength{\tabcolsep}{4.2pt}
\begin{tabular}{lccccc}
\toprule
Model & Prefill bs $1$ & Prefill bs $8$ & Decode bs $1$ & Decode bs $16$ & Decode bs $64$ \\
\midrule
Medium & $87.4\%$ ($148.9\textrm{k} / 170.3\textrm{k}$) & $90.4\%$ ($166.0\textrm{k} / 183.5\textrm{k}$) & $91.7\%$ ($565 / 616$) & $89.2\%$ ($2109 / 2365$) & $92.3\%$ ($2499 / 2709$) \\
Large  & $85.3\%$ ($95.3\textrm{k} / 111.7\textrm{k}$)  & $86.2\%$ ($104.0\textrm{k} / 120.6\textrm{k}$) & $85.4\%$ ($438 / 513$) & $87.8\%$ ($1916 / 2183$) & $84.9\%$ ($2311 / 2721$) \\
XL     & $83.8\%$ ($50.8\textrm{k} / 60.6\textrm{k}$)   & $85.3\%$ ($55.3\textrm{k} / 64.9\textrm{k}$)   & $86.9\%$ ($272 / 313$) & $84.6\%$ ($1245 / 1472$) & $86.6\%$ ($1452 / 1677$) \\
2.8B   & $87.5\%$ ($30.4\textrm{k} / 34.7\textrm{k}$)   & $90.5\%$ ($33.1\textrm{k} / 36.6\textrm{k}$)   & $86.4\%$ ($171 / 198$) & $87.6\%$ ($1248 / 1425$) & $88.9\%$ ($1836 / 2066$) \\
\bottomrule
\end{tabular}
\end{small}
\end{table}

\section{Extended Ablation Studies} \label{app:ablations}
\cref{tab:ablation-full} reports the complete 18-variant ablation suite at the Medium scale ($L = 18$, $D = 1200$, $n = 16$, $d = 75$, $d_k = 32$, $D_r = 256$, RoPE dim $\lfloor d/4 \rfloor = 18$, $T = 4096$, 7B tokens), extending \cref{tab:ablation}.
In full LoomFormer, $\op{FFN}_M$ has SwiGLU hidden width $1128$, and $\op{FFN}_x$ has width $2510$ in blocks $0$--$16$ and $3904$ in block $17$ (which omits unused writes to $M_L$).
Whenever an ablation removes parameters, we widen $\op{FFN}_x$ uniformly across blocks $0$--$16$ and absorb any embedding or integer-rounding residual in block $17$, matching the $432.11\textrm{M}$ parameter budget within $\pm 1.4\textrm{K}$ ($< 0.0004\%$).

\begin{table}[htb]
\centering
\caption{Complete Medium-scale (432M) ablation suite of LoomFormer under a matched parameter budget ($432.11\textrm{M} \pm 1.4\textrm{K}$).
$^{\dagger}$Included in \cref{tab:ablation}.}
\label{tab:ablation-full}
\begin{small}
\setlength{\tabcolsep}{3.7pt}
\begin{tabular}{llccc}
\toprule
Variant & Specification (vs.\ full LoomFormer) & Loss$\downarrow$ & $\Delta$Loss & Gain \\
\midrule
\B{LoomFormer} (full)$^{\dagger}$ & --- & \B{2.266} & --- & \B{100.0\%} \\
Transformer++ (baseline)$^{\dagger}$ & Standard MHA + SwiGLU FFN, no matrix stream $M$ & 2.443 & $+$0.177 & 0.0\% \\
\midrule
\multicolumn{5}{l}{\textit{Dynamic rewiring \& write-port gating (\cref{prop:ports})}} \\
\hspace{4pt}Static LoomMHA $k^m$, gated & Static $k^m$ in LoomMHA; write gate $\sigma(x W^W_g)$ kept & 2.266 & $0.000$ & 100.0\% \\
\hspace{4pt}Static LoomMHA $k^m$ & Static $k^m$ in LoomMHA; write gate dropped ($\equiv 1$) & 2.282 & $+$0.016 & 91.1\% \\
\hspace{4pt}Static $\op{FFN}_M$ $k^m$, gated & Static $k^m$ in $\op{FFN}_M$; write gate $\sigma(x W^W_g)$ kept & 2.268 & $+$0.001 & 99.3\% \\
\hspace{4pt}Static $\op{FFN}_M$ $k^m$ & Static $k^m$ in $\op{FFN}_M$; write gate dropped ($\equiv 1$) & 2.272 & $+$0.006 & 96.6\% \\
\hspace{4pt}Static write keys ($k^m$), gated & Static $k^m$ in emb., LoomMHA, $\op{FFN}_M$; write gates kept & 2.274 & $+$0.008 & 95.6\% \\
\hspace{4pt}Static write keys ($k^m$)$^{\dagger}$ & Static $k^m$ in emb., LoomMHA, $\op{FFN}_M$; write gates dropped & 2.283 & $+$0.017 & 90.3\% \\
\hspace{4pt}Static read queries ($q^m$)$^{\dagger}$ & Only static $\bar{q}$ in $\op{MRead}^{Q,K,V,O}$; all read gates dropped & 2.310 & $+$0.043 & 75.4\% \\
\hspace{4pt}Static read \& write ($q^m, k^m$), gated & Static $q^m$ (no read gates), all $k^m$; write gates kept & 2.327 & $+$0.060 & 65.8\% \\
\hspace{4pt}Static read \& write ($q^m, k^m$)$^{\dagger}$ & Static $q^m$, all $k^m$; port gates dropped ($\op{Tfm}_{K,Q}$ + $\op{MShift}$) & 2.344 & $+$0.078 & 56.0\% \\
\hspace{4pt}Static read \& write ($q^m, k^m$), $-$ $\op{MShift}$ & Static $q^m$, all $k^m$; port gates, $\op{MShift}$ dropped (Prop.~\ref{prop:ports}(i)) & 2.348 & $+$0.082 & 53.6\% \\
\midrule
\multicolumn{5}{l}{\textit{Port normalization \& gating (Eqs.~\ref{eq:MRead}--\ref{eq:MWrite})}} \\
\hspace{4pt}$-$ dynamic port gates$^{\dagger}$ & $\sigma(x W^R_g), \sigma(x W^W_g) \equiv 1$; $\sigma(x W^R_{\bar{g}})$ and $\op{MShift}$ kept & 2.277 & $+$0.011 & 93.7\% \\
\hspace{4pt}$-$ dynamic port $\op{Norm}$$^{\dagger}$ & Drop $\op{Norm}$ on dynamic $q^m, k^m$; $\op{Norm}(o_i)$, $\op{MNorm}$ kept & 2.289 & $+$0.023 & 87.1\% \\
\hspace{4pt}$-$ dynamic port gates \& $\op{Norm}$$^{\dagger}$ & Drop both $\sigma(x W^R_g), \sigma(x W^W_g)$ and port $\op{Norm}$ & 2.283 & $+$0.017 & 90.3\% \\
\midrule
\multicolumn{5}{l}{\textit{Sub-layer components (\cref{architecture})}} \\
\hspace{4pt}$-$ $\op{FFN}_M$ write$^{\dagger}$ & Remove $\op{FFN}_M$ and its $\op{MWrite}$ (Eq.~\ref{eq:loom-mlp}) & 2.300 & $+$0.034 & 80.7\% \\
\hspace{4pt}$-$ output bypass $o^m$$^{\dagger}$ & Drop $\op{MRead}^O$; restore self-attention to current token (Eq.~\ref{eq:loom-attn}) & 2.299 & $+$0.032 & 81.8\% \\
\hspace{4pt}$-$ vector-stream RoPE $W^{Q,K}$$^{\dagger}$ & Drop $W^{Q,K}$; RoPE on $18$ of $75$ dims of $\tilde{q}, \tilde{k}$ (Eq.~\ref{eq:loom-read}) & 2.286 & $+$0.020 & 88.7\% \\
\hspace{4pt}$-$ $\op{MShift}$$^{\dagger}$ & Set $z_{[t]} \equiv 0$ in Eq.~(\ref{eq:loom-read}) & 2.277 & $+$0.010 & 94.2\% \\
\hspace{4pt}$-$ LoomMHA write $\op{Norm}(o_i)$ & Drop $\op{Norm}$ on $o_i$ in LoomMHA writes (Eq.~\ref{eq:loom-attn}) & 2.268 & $+$0.002 & 98.8\% \\
\bottomrule
\end{tabular}
\end{small}
\end{table}

\textbf{Write-port gating vs.\ slot selection, and reduction to $\op{Tfm}_{K,Q}$.}
Fixing write keys $k^m$ to learned static vectors while keeping their dynamic gates $\sigma(x W^W_g)$ preserves nearly the full gain ($100.0\%$ for LoomMHA, $99.3\%$ for $\op{FFN}_M$, and $95.6\%$ across all three write sites including the embedding).
Dropping the gate on static $k^m$, however, hurts LoomMHA over three times as much as $\op{FFN}_M$ ($100.0\% \to 91.1\%$ vs.\ $99.3\% \to 96.6\%$): LoomMHA normalizes its head outputs $\op{Norm}(o_i)$ before writing (Eq.~\ref{eq:loom-attn}), leaving the gate as the sole control of write magnitude (and whether a head is on), whereas $\op{FFN}_M$ writes unnormalized slices (Eq.~\ref{eq:loom-mlp}) whose norms can still modulate write strength per token.
When both $q^m$ and $k^m$ are static, retaining dynamic write gates gives a node-gated circuit on fixed slots ($65.8\%$), dropping those gates leaves $\op{Tfm}_{K,Q}$ with $\op{MShift}$ ($56.0\%$), and further removing $\op{MShift}$ yields the exact $\op{Tfm}_{K,Q}$ of \cref{prop:ports}(i) ($53.6\%$)---which has no token-dependent port or shift gates on $M$, yet still outperforms all layer-wise dynamic baselines at the same scale (\cref{fig:scaling-loomformer}, where MUDDFormer reaches $48.5\%$ of LoomFormer's gain).

\textbf{Attention write-value normalization.}
Removing $\op{Norm}(o_i)$ from LoomMHA writes (Eq.~\ref{eq:loom-attn}) loses only $1.2\%$ of the gain ($98.8\%$ retained), as whole-matrix pre-normalization $\op{MNorm}$ before reads (Eq.~\ref{eq:loom-read}) already bounds the scale of $M$.

\section{Mechanistic Analysis Details and Visualizations} \label{interpretation}

\paragraph{Evaluation protocol and linear attribution.}
The statistics in \cref{analyzing} and this section are computed on held-out Pile validation sequences of length $T = 2048$ at the 300B-token checkpoints of Pythia-2.8B, mHC-Pythia-2.8B and LoomPythia-2.8B (evaluating token positions $t \ge 64$ to avoid sequence-boundary transients).
Because MShift and MNorm act linearly on the accumulated writes at each layer (\cref{eq:loom-read}), absorbing the per-slot shift gate $z_{[t]}$ (or $1 - z_{[t]}$) and the scalar MNorm factor into $c_{ij} = \langle q^m_j, k^m_i \rangle$ decomposes every MRead output as $u_j = \sum_i c_{ij} o_i$ over all earlier writers (embedding, LoomMHA and $\op{FFN}_M$ writes at $t$ and $t-1$; $0.4\%$ median RMS difference from BF16 execution across layers).
Projecting each term onto $u_j$ gives the additive writer attribution
\begin{equation} \label{eq:attribution}
    a_{ij} = \frac{c_{ij}\langle o_i, u_j\rangle}{\|u_j\|^2}, \qquad \textstyle\sum_i a_{ij} = 1.
\end{equation}

\paragraph{Definitions for \cref{fig:mech-main}.}
In \cref{fig:mech-main}(a), to isolate dynamic port rewiring from both static virtual weights (\cref{prop:ports}(i)) and per-token scalar gate or writer-norm fluctuations, let $\hat{q}^m_j(t) = q^m_j(t)/\|q^m_j(t)\|_2$ and $\hat{k}^m_i(t) = k^m_i(t)/\|k^m_i(t)\|_2 \in \mathbb{S}^{d_k-1}$ denote the unit port directions of reader $j$ and same-token writer $i$, weighted by each writer's corpus RMS norm $\bar{\omega}_i = \sqrt{\mathbb{E}_t[\|o_i(t)\|^2]}$: $\widehat{C}_{ij}(t) = \bar{\omega}_i \langle \hat{q}^m_j(t), \hat{k}^m_i(t)\rangle$.
We center across tokens, $\Delta\widehat{C}(t) = \widehat{C}(t) - \mathbb{E}_{t'}[\widehat{C}(t')]$, and report the energy effective rank $\exp\!\big({-}\sum_k p_k \log p_k\big)$ with $p_k = \sigma_k^2 / \sum_{k'} \sigma_{k'}^2$ over the $3n = 96$ Q/K/V readers of layer $l$ (or the $n = 32$ readers of a single Q, K or V input); under fixed-port $\op{Tfm}_{K,Q}$ with arbitrary positive scalar gates, $\Delta\widehat{C}(t) \equiv 0$.
In \cref{fig:mech-main}(b), for each read port $q^m_j(t) \in \R^{d_k}$ ($\tilde{q}, \tilde{k}, v, o^m$) and write port $k^m_i(t) \in \R^{d_k}$ (LoomMHA and $\op{FFN}_M$), we average $1 - \cos(p(t_1), p(t_2))$ over distinct token pairs $t_1 \ne t_2$ ($t \ge 64$) and over the $n = 32$ heads of layer $l$ (embedding write ports remain near-static at $0.005$, providing a sanity check).
In \cref{fig:mech-main}(c), following \citet{xiao2025muddformer}, a head is on at query position $t$ if $\arg\max_s \alpha_{ts}$ falls neither in $\{0, 1\}$ nor on a sink token (\texttt{<|endoftext|>}, ``\texttt{.}'', ``\texttt{\ .}'', ``\texttt{\textbackslash n}'', ``\texttt{\textbackslash n\textbackslash n}''); in layers 17--32, first-two-token sink mass drops from $26.4\%$ (Pythia) and $27.2\%$ (mHC) to $1.5\%$ in LoomPythia while content-token attention mass rises from $22.7\%$ and $31.2\%$ to $65.4\%$, and masking the self-diagonal $s = t$ in Pythia and mHC to match LoomMHA's strict causal mask (\cref{eq:loom-attn}) leaves the gap unchanged ($74.4\%$ vs.\ $21.2\%$ and $15.7\%$).
In \cref{fig:mech-main}(d), for each head $h$ in layers 3--32 and each pair among its four ports $\{\tilde{q}_h, \tilde{k}_h, v_h, o^m_h\}$, we compute from token-averaged positive attributions $\bar{a}^+_{ij} = \mathbb{E}_t[\max(a_{ij}(t), 0)]$ over same-token writers: (i) Pearson correlation after summing attributions by writer group (Emb, each earlier LoomMHA layer, each earlier $\op{FFN}_M$ layer), (ii) Pearson correlation across individual head/module writers, and (iii) $|S_A \cap S_B|/32$ for the top-32 writer sets $S_A, S_B$, reporting the median across heads (in slot space $\R^{d_k}$, same-token cosine similarity between ports of the same head averages $0.01$--$0.07$ outside Q--K and $0.27$ on Q--K).

\begin{figure}[htb]
\centering
\includegraphics[width=0.88\textwidth]{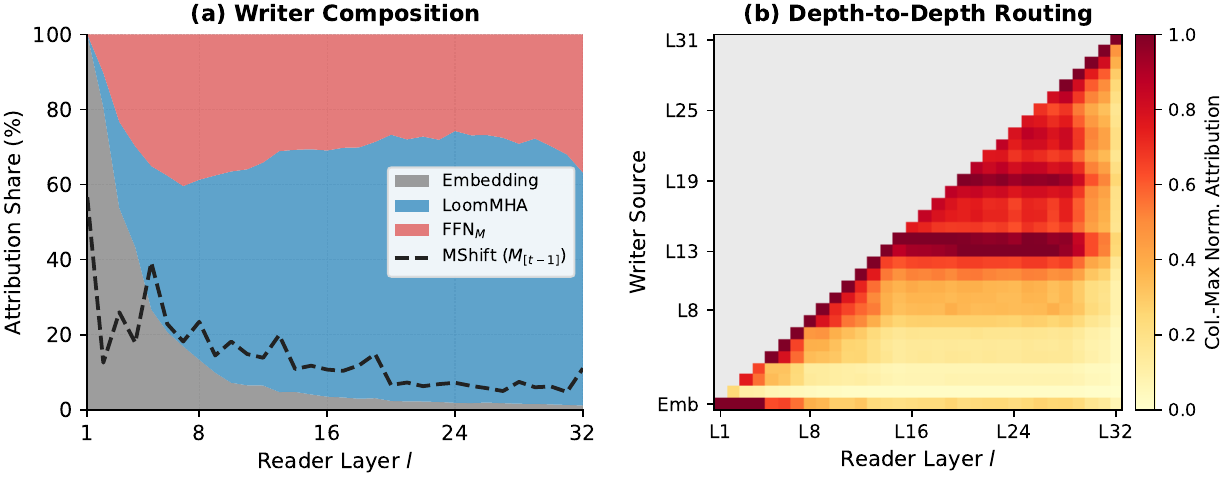}
\vskip -0.08in
\caption{Matrix-stream routing in LoomPythia-2.8B across Pile validation sequences.
(a) Share of positive attribution from Embedding, LoomMHA and $\op{FFN}_M$ writes, and share of $|a_{ij}|$ from previous-token MShift ($M_{[t-1]}$).
(b) Column-max normalized positive attribution from each writer source (vertical, bottom to top) to each reader layer $l$ (horizontal; gray upper triangle: causally unreachable).}
\label{fig:mech-routing}
\vskip -0.08in
\end{figure}

\paragraph{Writer composition and depth-to-depth routing.}
Averaged over layers 2--32 in LoomPythia-2.8B, positive attribution mass $\max(a_{ij}, 0)$ is shared across LoomMHA writes ($53.4\%$--$62.5\%$ across $\tilde{q}, \tilde{k}, v, o^m$), $\op{FFN}_M$ writes ($27.5\%$--$35.2\%$) and embedding writes ($10.0\%$--$11.7\%$), while previous-token MShift reads account for $13.7\%$--$15.2\%$ of $\sum_i |a_{ij}|$ (\cref{fig:mech-routing}(a)).
$\op{FFN}_M$ reaches its largest share ($35.2\%$) on value reads $v$, consistent with $\op{FFN}_M$ serving as the sole post-embedding channel from the vector stream into $M$ (\cref{eq:loom-mlp}) and with its large effect in \cref{tab:ablation}.
Plotted with writer sources on the vertical axis (bottom to top) and reader layers on the horizontal axis, the column-max normalized routing map (\cref{fig:mech-routing}(b)) combines local relay along the diagonal with persistent horizontal bands from hub writer sources---most prominently the embedding at early layers and writer layers 13--14 and 19 across all subsequent layers.

\begin{figure}[htb]
\centering
\includegraphics[width=\textwidth]{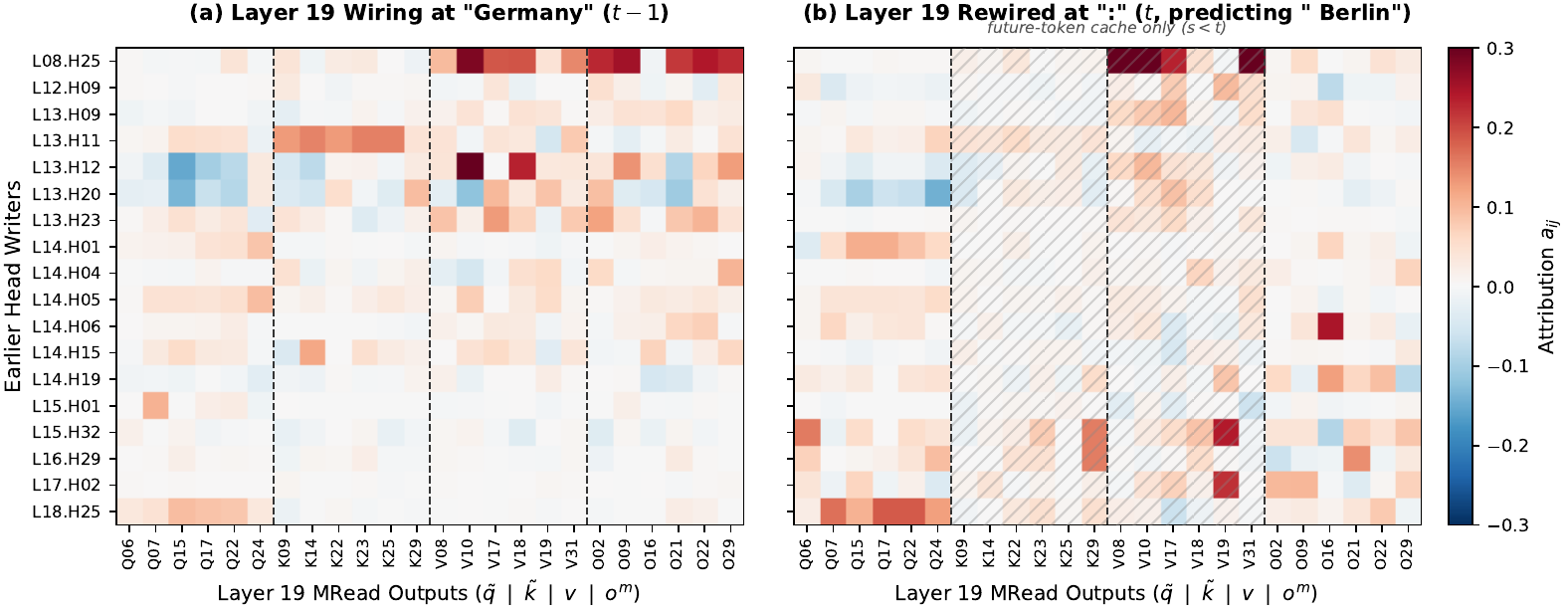}
\vskip -0.08in
\caption{Attribution submatrix $[a_{ij}]$ (rows: top-18 earlier head writers \texttt{L}$l$\texttt{.H}$h$; columns: top-6 layer-19 MRead outputs per arm $\tilde{q}, \tilde{k}, v, o^m$) of LoomPythia-2.8B at two consecutive tokens of an in-context prompt (next-token top-1: \texttt{" Berlin"}, $p = 0.65$), showing non-rectangular within-token wiring at \texttt{"Germany"} (a) and per-token rewiring onto causal $\tilde{q}$ and $o^m$ pathways at \texttt{":"} (b; hatched $\tilde{k}, v$ columns affect only future tokens under $s < t$).}
\label{fig:mech-icl}
\vskip -0.1in
\end{figure}

\paragraph{Adjacent-token rewiring on an in-context prompt.}
\cref{fig:mech-icl} shows the head-writer attribution submatrix $[a_{ij}]$ at hub layer 19 (selecting the top 6 readers per MRead arm and top 18 earlier LoomMHA head writers by peak $|a_{ij}|$ across the two tokens) on \texttt{"France: Paris\textbackslash nJapan: Tokyo\textbackslash nChina: Beijing\textbackslash nSpain: Madrid\textbackslash nGermany:"}, where LoomPythia-2.8B predicts \texttt{" Berlin"} top-1 with probability $0.65$.
At \texttt{"Germany"} (\cref{fig:mech-icl}(a)), writers route selectively to distinct MRead arms---\texttt{L13.H11} feeds key reads ($\tilde{k}$: \texttt{K14}, \texttt{K23}, \texttt{K25}) without feeding value or bypass reads, whereas \texttt{L13.H12} and \texttt{L08.H25} feed value and bypass reads (\texttt{V10}, \texttt{O09}, \texttt{O22})---forming a non-rectangular within-token pattern (\cref{prop:rectangle}) that prepares \texttt{"Germany"}'s horizontal keys, values and residual state.
At the immediately following delimiter \texttt{":"} (\cref{fig:mech-icl}(b)), LoomMHA's strict causal mask ($s < t$, \cref{eq:loom-attn}) means that only $\tilde{q}$ (attending back to \texttt{"Germany"} and the few-shot capitals) and $o^m$ affect the \texttt{" Berlin"} prediction, whereas $\tilde{k}$ and $v$ at \texttt{":"} serve only future tokens.
Correspondingly, \texttt{L13.H11} and \texttt{L13.H12} switch off on layer 19, while layer 19 rewires its causal query reads $\tilde{q}$ onto \texttt{L18.H25}, \texttt{L15.H32} and \texttt{L14.H01} (\texttt{Q06}, \texttt{Q07}, \texttt{Q17}, \texttt{Q22}) and its bypass outputs $o^m$ onto \texttt{L14.H06}, \texttt{L16.H29} and \texttt{L14.H19} (\texttt{O16}, \texttt{O21}); causally freezing layer 19's dynamic read ports at \texttt{":"} to those at \texttt{"Germany"} alone drops $P(\text{\texttt{" Berlin"}})$ from $0.65$ to $0.54$ ($0.41$ when frozen across all layers).

\end{document}